\documentclass[lettersize,journal]{IEEEtran}
\usepackage{amsmath,amsfonts}
\usepackage{algorithmic}
\usepackage{algorithm}
\usepackage{array}
\usepackage[caption=false,font=normalsize,labelfont=sf,textfont=sf]{subfig}
\usepackage{textcomp}
\usepackage{stfloats}
\usepackage{url}
\usepackage{verbatim}
\usepackage{graphicx}
\usepackage{cite}
\usepackage{multirow}
\usepackage{hyperref}
\usepackage[table]{xcolor}
\usepackage{booktabs}

\usepackage{soul}

\begin{document}

\title{SAGD: Boundary-Enhanced Segment Anything in 3D Gaussian via Gaussian Decomposition}

\author{Xu Hu$^{1,2}$,
        Yuxi Wang$^{2,3}$,
        Lue Fan$^{3,4}$,
        Chuanchen Luo$^{6}$,
        Junsong Fan$^{2,3}$,
        Zhen Lei$^{2,3,4}$, \\
        Qing Li$^{1\textsuperscript{†}}$,
        Junran Peng$^{5}$\textsuperscript{†},
        and Zhaoxiang Zhang$^{2,3,4}$\textsuperscript{†}

        \thanks{\IEEEcompsocthanksitem \textsuperscript{†} Corresponding author.
\IEEEcompsocthanksitem $^{1}$ The Hong Kong Polytechnic University 
\IEEEcompsocthanksitem $^{2}$ Center for Artificial Intelligence and Robotics, HKISI, CAS
\IEEEcompsocthanksitem $^{3}$ Institute of Automation, Chinese Academy of Sciences
\IEEEcompsocthanksitem $^{4}$ University of Chinese Academy of Sciences
\IEEEcompsocthanksitem $^{5}$ University of Science and Technology Beijing
\IEEEcompsocthanksitem $^{6}$ Shandong University
}
}


\maketitle

\begin{abstract}









3D Gaussian Splatting has emerged as an alternative 3D representation for novel view synthesis, benefiting from its high-quality rendering results and real-time rendering speed. 
However, the 3D Gaussians learned by 3D-GS have ambiguous structures without any geometry constraints. This inherent issue in 3D-GS leads to a rough boundary when segmenting individual objects. 
To remedy these problems, we propose SAGD, a conceptually simple yet effective boundary-enhanced segmentation pipeline for 3D-GS to improve segmentation accuracy while preserving segmentation speed.
Specifically, we introduce a Gaussian Decomposition scheme, which ingeniously utilizes the special structure of 3D Gaussian, finds out, and then decomposes the boundary Gaussians. 
Moreover, to achieve fast interactive 3D segmentation, we introduce a novel training-free pipeline by lifting a 2D foundation model to 3D-GS.
Extensive experiments demonstrate that our approach achieves high-quality 3D segmentation without rough boundary issues, which can be easily applied to other scene editing tasks. 
Our code is publicly available at \url{https://github.com/XuHu0529/SAGS}.

\end{abstract}

\begin{IEEEkeywords}
3D gaussian splatting, 3D segmentation, boundary issues.
\end{IEEEkeywords}

\section{Introduction}
\label{sec:intro}


3D scene understanding is a challenging and crucial task in computer vision and computer graphics, which involves scene reconstruction from images or videos and the perception of a given 3D real-world environment. Researchers have conducted extensive studies in scene reconstruction and 3D scene perception in recent years. For instance, Neural Radiance Fields (NeRF) \cite{Nerf, TensorRF, TensorIR,DVGO} have significantly contributed to the progress of 3D scene reconstruction by representing scenes in an implicit way. 
In the field of scene perception, continuous research is being conducted on 3D detection and semantic segmentation, based on the representation of range images \cite{LaserNet,RangeDet}, point clouds \cite{lai2022stratified,FSD,PointFormer,PointRCNN,hu2021towards}, and Bird's Eye View (BEV) \cite{BEAFormer,BEVFormerv2,BAEFormer,BEV-Seg,BEVSegFormer}. Although current methods have attained noteworthy success in 3D scene understanding, the time-consuming nature of NeRF and the high costs associated with 3D data collection pose challenges for scaling up these approaches.

\begin{figure*}[!tbp]
    \centering    \includegraphics[width=0.98\linewidth]
    {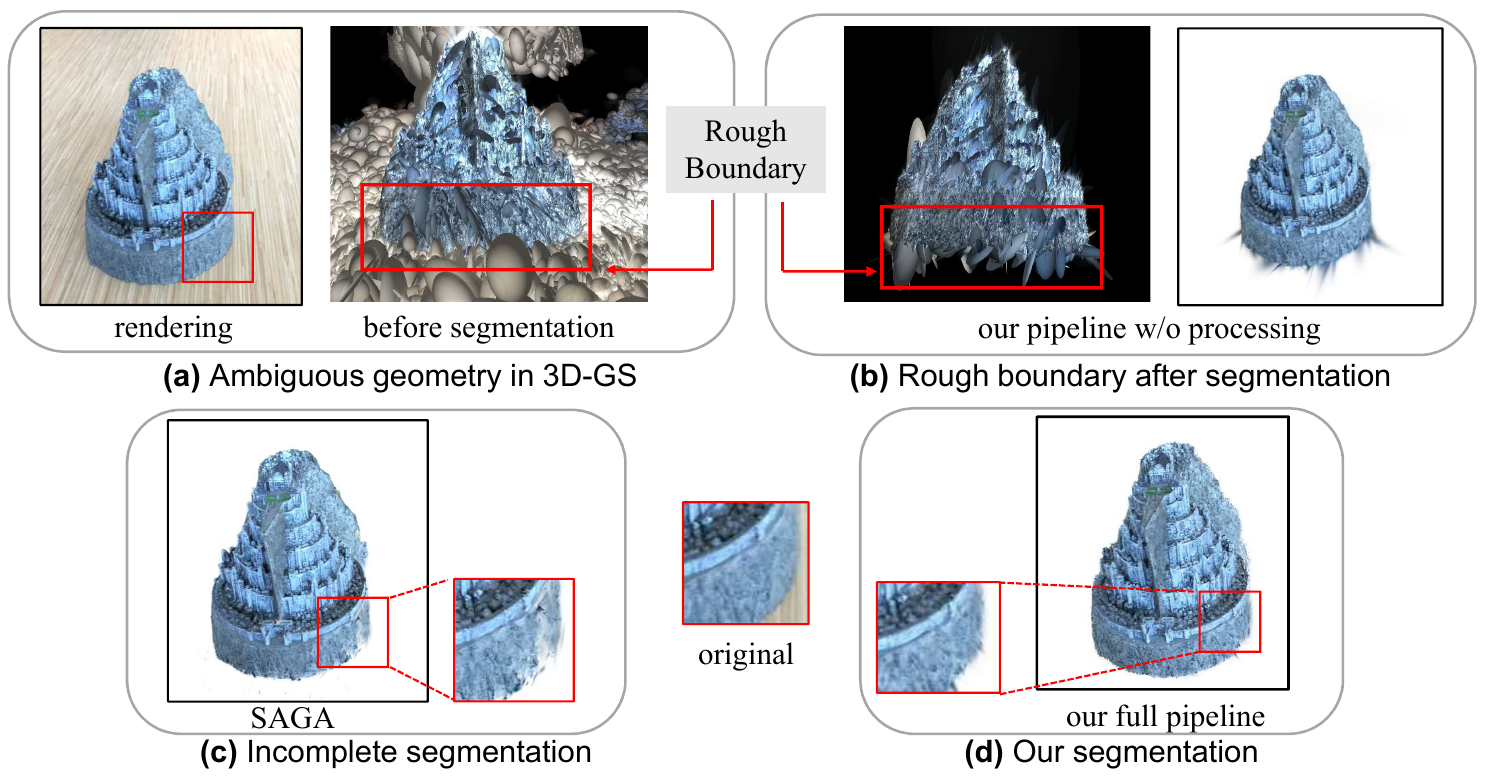}
    \caption{(a)The training of 3D-GS doesn't consider the structure of objects, leading to the ambiguous geometry; (b)Direct segmentation without Gaussian Decomposition processing will result in rough boundary segmentation; (3) The recent SAGA also has incomplete segmentation caused by the same issue; (d) Our full pipeline considers this issue and can achieve better segmentation.}
    \label{fig:intro}
    \vspace{-0.5cm}
\end{figure*}

Recently, 3D Gaussian Splatting (3D-GS) \cite{3DGS} is emerging as a prospective method for modeling static 3D scenes. 3D-GS characterizes intricate scenes by employing numerous colored 3D Gaussians, rendering them into camera views through splatting-based rasterization. Through differentiable rendering and gradient-based optimization, the positions, sizes, rotations, colors, and opacities of these Gaussians can be finely tuned to accurately represent the 3D scene, availing for the comprehension of a 3D environment. 
Segmentation in 3D-GS is quite a natural pathway to scene understanding for its explicit representation. Recent works~\cite{SAGA, gaussian-grouping, featureGS, 2d-guided} have been proposed to achieve segmentation in 3D-GS via lifting 2D foundation models in a learnable way. They share the same ideas to distill other features or identity encodings generated from 2D foundation models to 3D Gaussian fields.

However, the Gaussians learned by 3D-GS are ambiguous without any geometry constraint, leading to the issue of boundary roughness. 
A single Gaussian might correspond to multiple objects, complicating the task of accurately segmenting individual objects (shown in Fig.~\ref{fig:intro} (a)).  
As shown in Fig.~\ref{fig:intro} (b), direct segmentation will leave rough edges at the boundary. This is because these Gaussians across multiple objects will be left if no additional processing is performed. In addition, even if SAGA~\cite{SAGA} uses filtering and growing post-processing for segmented 3D Gaussians, its feature matching method also has the limitation of incomplete boundary, as shown in Fig.~\ref{fig:intro} (c).

We present our boundary-enhanced segmentation method, an interactive training-free pipeline for efficient and effective segmentation in 3D-GS without rough boundary issues.
By leveraging the 2D foundation model SAM~\cite{SAM}, our method can generate a segmented mask from a single input view based on the given prompts. Starting from the obtained mask, our method automatically generates multi-view masks and achieves consistent 3D segmentation via the proposed assignment strategy. 
Specifically, to resolve the inherent boundary issues, we incorporate a simple but effective Gaussian Decomposition scheme, ingeniously utilizing the special structure of 3D Gaussian. 
Since the boundary roughness issue results from the non-negligible spatial sizes of 3D Gaussians located at the boundary, our key insight is to find and then decompose the original boundary Gaussians according to our proposed principles. 
Consequently, our method eliminates the inherent issues and achieves more complete segmentation quickly and efficiently, as shown in Fig.~\ref{fig:intro} (d).

In summary, the contributions of this paper are as follows: 
\begin{itemize}
    \item We propose a simple yet effective training-free pipeline for segmentation in 3D Gaussians without any learnable parameters;
    \item We incorporate the Gaussian Decomposition module to mitigate the boundary roughness issues in 3D segmentation resulting from inherent 3D-GS geometry;
    \item Extensive segmentation experiments on a considerable amount of 3D scenes and editing applications demonstrate the effectiveness of our proposed method.
    
\end{itemize}

\vspace{-0.3cm}
\section{Related Work}
\label{sec:related-work}

\subsection{Radiance Fields and 3D Gaussian Splatting}
Novel view synthesis (NVS) involves rendering unseen viewpoints of a scene from a given set of images. One popular approach is Neural Radiance Fields (NeRF), which uses a Multilayer Perceptron (MLP) to represent 3D scenes and map from a 3D coordinate to properties of the scene at the corresponding location. It leverages the differentiable volume rendering technique to translate a 3D scene's continuous representation into 2D images.
Several works have been proposed to enhance NeRF’s performance by addressing aspects such as speed~\cite{DVGO,dvgov2,TensorRF,InstantNGP, plenoxel, fastnerf, kilonerf, plenoctrees, donerf} and adapting it to other tasks~\cite{a-nerf,relighting, humannerf, nerfactor}.
Though some breakthroughs have been achieved, the reliance on low-efficient volume rendering still hinders real-time rendering.

Recently, 3D Gaussian Splatting~\cite{3DGS} has been proposed as a new technique for novel view synthesis. 
It leverages an ensemble of anisotropic 3D Gaussian splats to represent the scene and employs differentiable splatting for rendering.
3D Gaussian Splatting has been shown to be an alternative 3D representation of NeRF, benefiting from both its high-quality rendering results and real-time rendering speed. 
Recent research on this technique involves applying 3D Gaussian Splatting to large-scale scene reconstruction~\cite{liu2025citygaussian, liu2024citygaussianv2}, the dynamic scenes~\cite{luiten2023dynamic, wu20244d, ling2024align, zhou2024drivinggaussian, zeng2024stag4d, bae2024per,lu20243d, shao2024control4d, huang2024sc, lin2024gaussian, yu2024cogs,li2024spacetime,sun20243dgstream}, 3D object and scene generation~\cite{tang2023dreamgaussian, yi2023gaussiandreamer,chen2024text, xu2024agg, yang2024gaussianobject, zhou2024dreamscene360, liang2024luciddreamer, zhang2024cityx, zhou2024scenex}, 
surface reconstruction~\cite{huang20242d, gof, surfels, guedon2024sugar} and other tasks~\cite{charatan2024pixelsplat, wewer2024latentsplat, liang2024gs, shao2024control4d}.


\subsection{2D and 3D Perception}
Detection and segmentation are fundamental computer vision tasks.
Numerous studies~\cite{peng2018accelerating,chang2022data,bu2021gaia,peng2019efficient,peng2019pod,zhang2022delving,peng2020large,zhang2025general,peng2023gaia,zhang2024voxel} have deeply explored various sub-fields. Especially, Grounding-DINO~\cite{groundingdino} enhances the grounding results by introducing the language instructions.
Traditionally, the segmentation includes three major tasks: semantic~\cite{chen2014semantic,chen2017rethinking,chen2017deeplab,chen2018encoder,wang2021uncertainty}, instance~\cite{hafiz2020survey,liu2018path,bolya2019yolact}, and panoptic~\cite{Kirillov_2019_CVPR} segmentation. 
Due to the operation consistency of the above tasks at the pixel level, many studies have tried to use a unified framework, such as K-net~\cite{zhang2021knet}, MaskFormer~\cite{cheng2021maskformer}, and Mask2Former~\cite{cheng2022maskedattention}. A significant breakthrough in 2D segmentation is the Segment Anything Model (SAM)~\cite{SAM}. SAM seeks to unify the 2D segmentation task using a prompt-based segmentation approach. Its introduction has sparked a new wave of research, with many studies focused on enhancing its functionality. 
These improvements include efficient fine-tuning techniques~\cite{tunesam1,tunesam2,hqsam} and distillation-based acceleration methods~\cite{fastsam,mobileSAM}. 
Additionally, SAM has been adapted for various fields, such as medical image analysis~\cite{medicalSAM, mazurowski2023segment, wu2023medical, deng2023segment}, concealed object detection~\cite{ConcealedSAM, tang2023can}, image editing~\cite{yu2023inpaint}, remote sensing~\cite{chen2024rsprompter}, and 3D bird's eye view (BEV) sensing~\cite{zhang2023sam3d}. 

\subsection{Segmentation in Radiance Fields}

Neural Radiance Fields (NeRFs) are a popular way of implicitly representing 3D scenes with neural networks. Many researchers have explored how to segment objects in 3D using NeRFs for various applications such as novel view synthesis, semantic segmentation, 3D inpainting, and language grounding. Some methods, such as Semantic-NeRF~\cite{semantic-nerf}, NVOS~\cite{NVOS}, and SA3D~\cite{SA3D}, use different types of inputs to guide the segmentation, such as semantic labels, user scribbles, or 2D masks. Other methods, such as N3F~\cite{N3F}, DFF~\cite{N3F}, LERF~\cite{lerf}, and ISRF~\cite{isrf}, learn additional feature fields that are aligned with NeRFs, and use 2D visual features from pre-trained models or language embeddings to query the 3D features. These methods usually require modifying or retraining the original NeRF models or training other specific parameters to obtain the 3D segmentation. However, limited by the representation and rendering speed of NeRF, it is challenging to apply it to more practical applications, such as scene editing, collision analysis, etc. 
Inspired by the real-time 3D-GS, recent works~\cite{SAGA, gaussian-grouping, featureGS, 2d-guided} have been proposed to achieve segmentation in 3D-GS in a learnable way. SAGA~\cite{SAGA} and Featrue 3DGS~\cite{featureGS} distill the knowledge embedded in the SAM decoder into the feature field of 3D-GS, and Gaussian-Grouping~\cite{gaussian-grouping} supervises the Identity Encodings during the differentiable rendering by leveraging the 2D mask predictions by SAM. 
However, no research has been proposed to solve the rough boundary issues in segmentation resulting from the inherent properties of 3D Gaussians. In this work, we are the first to propose a novel Gaussian Decomposition scheme to solve this issue, incorporated into our effective training-free segmentation pipeline.
\newcommand{\g}{\mathbf{g}}
\newcommand{\p}{\mathbf{p}}
\newcommand{\m}{\mathbf{m}}
\newcommand{\bv}{\mathbf{v}}
\newcommand{\bc}{\mathbf{c}}


\begin{figure*}[!tbp]
    \centering
    \includegraphics[width=0.99\textwidth]{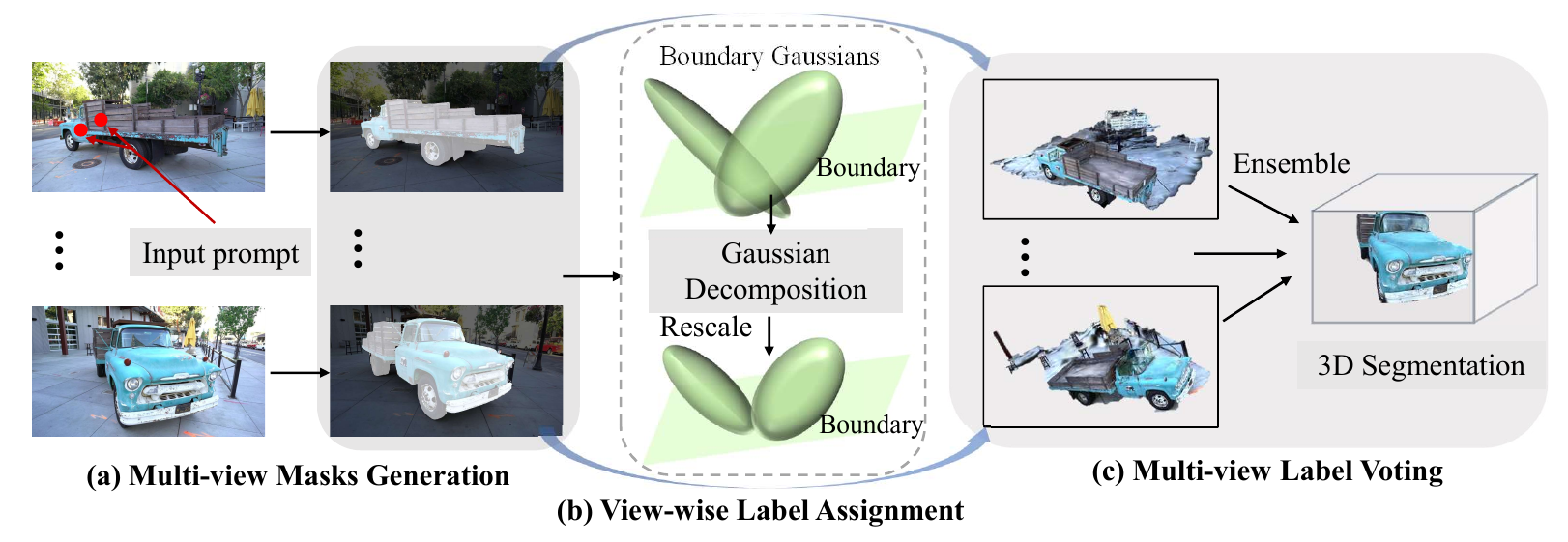}
    \caption{Pipeline of our proposed method. (a) Given a set of clicked points on the $1^{st}$ reference view, we utilize SAM to generate masks for corresponding objects under every view automatically; (b) For every view, Gaussian Decomposition is first performed to address the issue of boundary roughness and then label propagation is implemented to assign binary labels to each 3D Gaussian; (c) Finally, with assigned 3D labels from all views, we adopt a simple yet effective voting strategy to determine the segmented Gaussians.}
    \label{fig:pipeline}
    \vspace{-0.5cm}
\end{figure*}

\section{Method} \label{method}

In this section, we first present the preliminary in 3D Gaussian Splatting and Segment Anything Model (SAM) in Sec.~\ref{sec:prelinminary} for clear understanding and then define the problem and task in Sec.~\ref{sec:definition}.
Our basic training-free pipeline consists of Sec.~\ref{sec:mask_assign} and Sec.~\ref{sec:vote}.
The details of Gaussian Decomposition are described in Sec.~\ref{sec:decomp}.
The pipeline of our method can be seen in Fig.~\ref{fig:pipeline}.

\subsection{Preliminary}
\label{sec:prelinminary}

\paragraph{3D Gaussian Splatting}
3D Gaussian Splatting (3D GS)~\cite{3DGS} is an emerging method for real-time radiance field rendering.
It has been proven effective in Novel View Synthesis with high rendering quality as NeRF and real-time rendering speed. 
3D GS represents scenes with a set of 3D Gaussians.
Specifically, each 3D Gaussian is parameterized by a position $\mu \in \mathbb{R}^3$, a covariance matrix $\Sigma$ consisting of a scaling factor $s \in \mathbb{R}^3$, and a rotation quaternion $q \in \mathbb{R}^4$, an opacity value $\alpha$, and spherical harmonics (SH).
Each 3D Gaussian is characterized by:
\begin{equation}
    G(x)=\frac{1}{\left(2\pi\right)^{3/2}\left|\Sigma\right|^{1/2}}e^{-\frac{1}{2}{\left(x - \mu\right)}^T\Sigma^{-1}\left(x - \mu\right)}
\label{eq:3dgs}
\end{equation}
\par
To render an image, it uses the splatting rendering pipeline, where 3D Gaussians are projected onto the 2D image plane. 
The projection transforms 3D Gaussians into 2D Gaussians in the image plane.
All 2D Gaussians are blended together by the $\alpha$-blending algorithm to generate the color:
\begin{equation}
    \bc = \sum_{i \in N} \bc_i\alpha_i \prod_{j=1}^{i-1} (1-\alpha_j),
\end{equation}
During the $\alpha$-blending process, for each 2D Gaussian, only the 2D points with probability density larger than a certain threshold are calculated.
This means a 2D Gaussian and 3D Gaussian can be intuitively regarded as a 2D \textbf{ellipse} and a 3D \textbf{ellipsoid} respectively.
Empirically, for an axis of the ellipse, its length is set to $3\sigma$, where $\sigma$ is the square root the of variance in the axis.


\paragraph{Segment Anything Model (SAM)} SAM\cite{SAM} takes an image $I$ and a set of prompts $P$ as input to output the corresponding segmentation mask $M_{SAM}$:

\begin{equation}
    M_{SAM} = SAM(I, P)
\end{equation}
where the prompts $P$ can be points, bounding boxes, or texts.


\subsection{Problem Definition}
\label{sec:definition}
Given a set of trained 3D gaussians $\mathbb{G} = \{\g_0, \g_1, \cdots, \g_n\}$ and a random initial view $\mathbf{v}_0$, users could offer 2D point prompt set $\mathbb{P}_{2D} = \{\p_0, \p_1, \cdots, \p_m\}$ to specify a 2D object in view $\mathbf{v}_0$.
Our algorithm is supposed to segment the corresponding 3D object $\mathbb{O}$ in $\mathbb{G}$ according to the human prompts, where $\mathbb{O}$ is a subset of  $\mathbb{G}$.
\par
Let $\m_i$ denote the projected binary mask of $\mathbb{O}$ in $i$-th view, an accurate segmentation $\mathbb{O}$ means
$\m_i$ equals to $\m_i^*$ for $\forall i \in \{0, 1, \cdots, n\}$.
Here $\m_i^*$ is the ground truth mask of $\mathbb{O}$ in  $i$-th view.
Different from the conventional 2D segmentation mask in the image or 3D segmentation task in the point cloud, there is no ground truth for 3D Gaussians.
Thus, our algorithm is designed to minimize the difference between $\m_i$ and $\m_i^*$.

\subsection{Segment 3D Gaussians with 2D Mask}
\label{sec:mask_assign}

\noindent \textbf{\textit{3D Prompts for Multiview Masks Generation.}}
By the definition in Sec.~\ref{sec:definition}, users are given the first rendered view to specify the target object.
However, only the first view is far from sufficient to segment the target object in 3D space. 
So here we first generate multiview masks to aid the 3D segmentation.
With multiple masks from different views, a 3D object can be segmented by the intersection of the corresponding frustum of these masks.
\par
The core of obtaining masks is to generate 2D prompt points in each view. 
Denoting the $i$-th 2D prompt point in the first given view $\mathbf{v}_0$ as $\p_i^0$, we define the corresponding 3D prompt $\p_i^{3D}$ as:
\begin{equation}
   \arg\min_{\mu}\{d(\mu), d(\mu) > 0 \mid \mu \in \mathbb{G},  \|\mathbf{P}_0\mu - \p_i^0\|_1 < \epsilon  \},
   \label{eq:prompt_3d}
\end{equation}

where $d(\mu)$ is the depth of  Gaussian center $\mu$ and $\mathbf{P}_0$ is the projection for initial view $\bv_0$.
Thus $\mathbf{P}_0\mu$ is the position of $\mu$ in view $\bv_0$.
Eq.~\ref{eq:prompt_3d} indicates that the corresponding 3D prompt of $\p_i$ is the center of a certain 3D Gaussian. This center meets two requirements: (1) it has a similar projected position with $\p_i$ with a Manhattan distance less than $\epsilon$, and (2) if there are multiple 3D Gaussian centers satisfying the first requirement, the one with the smallest positive depth is selected as the 3D prompt.

For all the 2D prompt points in the first view, we could get a set of 3D prompts by Eq.~\ref{eq:prompt_3d}.
Then for another view $\bv_i$, we project these 3D prompts into the 2D plane, resulting in 2D prompts in the view $\bv_i$.
In this way, we obtain 2D prompts in all views, and all masks are obtained by prompting the SAM.

\noindent \textbf{\textit{View-wise Label Assignment.}}
With all the masks, we proceed to assign binary labels to each 3D Gaussian.
In particular, we maintain a matrix $\mathbf{L}$, whose element $\mathbf{L}_{ij}$ is defined by

\begin{equation}
\mathbf{L}_{ij} =
    \begin{cases}
        1 & \text{if } \mathbf{P}_j\mu_i \in \m_j, \\
        0 & \text{if } \mathbf{P}_j\mu_i \notin \m_j,
    \end{cases}
    \label{eq:mask}
\end{equation}

where $\mu_i$ is the $i$-th Gausssian center of the scene and $\m_j$ is the foreground mask in $j$-th view.
$\mathbf{P}_j$ stands for the projection matrix of $j$-th view.

\begin{figure}[!t]
    \centering    \includegraphics[width=0.99\linewidth]{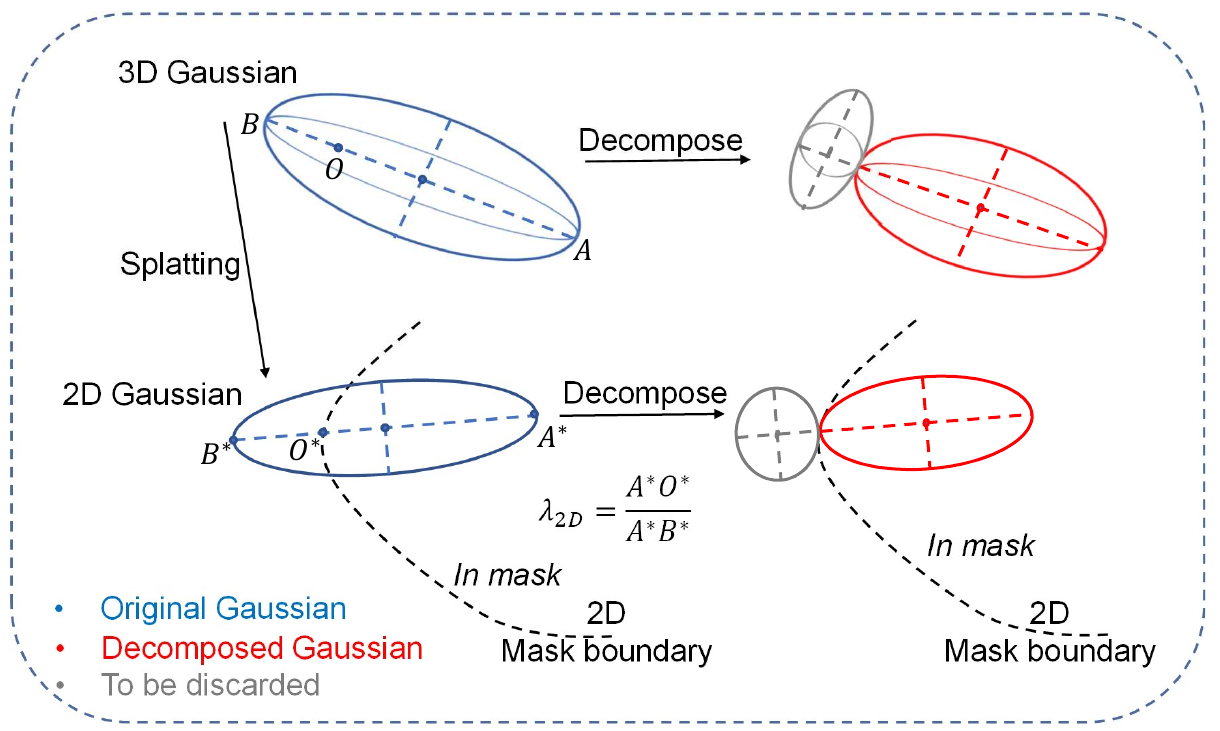}
    \caption{Illustration of the Gaussian Decomposition process. It involves two basic steps: first, to find out the boundary Gaussians and then decompose these Gaussians.}
    \label{fig:gd}
    \vspace{-0.5cm}
\end{figure}

\subsection{Gaussian Decomposition}
\label{sec:decomp}
After obtaining the prompt points of each view, we could assign labels to 3D Gaussians by the projected position of its center.
However, 3D Gaussian has non-negligible spatial volume, so those Gaussians projected to the mask boundary usually have a part out of the boundary, greatly increasing the roughness of boundaries.
A straightforward solution is directly removing the Gaussians across mask boundaries.
However, such a solution greatly damages the 3D structures of the object.

To address this issue, we propose the \emph{Gaussian Decomposition} to mitigate the boundary roughness while maintaining the 3D structures as complete as possible.
Fig.~\ref{fig:gd} illustrates the basic idea of Gaussian decomposition.
It has two basic steps to achieve the Gaussian decomposition:

\noindent
\textbf{(1)} For each 3D Gaussian, we obtain the corresponding 2D Gaussian by projection and mark it as a boundary Gaussian if one of its long-axis endpoints is outside of the 2D mask.
Then, as shown in Fig.~\ref{fig:gd}, we denote the two ends of the long axis of a 2D Gaussian as $A^*$ and $B^*$, and the intersection of $A^* B^*$ and mask boundary is $O^*$. $A$, $B$, and $O$ are the corresponding 3D points in the long axis of the 3D Gaussian.
Assuming $A^*$ is in the mask while $B^*$ is out of the mask, we define 
\begin{gather}
\lambda_{3D} = \frac{OA}{AB}, \\
     \lambda_{2D} = \frac{O^* A^*}{A^* B^*}.
\end{gather}
\textbf{(2)} After obtaining the boundary Gaussians, we solve the ratio $\lambda_{3D}$ by first calculating $\lambda_{2D}$ and then decompose the original 3D Gaussian according to $\lambda_{3D}$.
\par
However, the transformation from $\lambda_{2D}$ to $\lambda_{3D}$ is not straightforward because the perspective projection from 3D space to 2D space is not affine, which means the ratios are different.
Fortunately, 3D Gaussian Splatting leverages local affine approximation to simplify the rendering process. It projects 3D Gaussian to the 2D plane by
\begin{equation}
    \Sigma^\prime = JW\Sigma W^TJ^T
    \label{eq:affine}
\end{equation}
where $\Sigma$, $\Sigma^{\prime}$ are the covariance matrix of the 3D and 2D gaussian distribution, respectively. $W$ is the projection transformation and $J$ is the Jocobian of affine approximation derived in EWA algorithm~\cite{ewa}.
Eq.~\ref{eq:affine} indicates that Gaussian Splatting simplifies the perspective projection to an affine projection, thus decomposition scaling ratio $\lambda_{3D}$ in 3D space is equivalent to the ratio $\lambda_{2D}$ in the 2D plane.

Let $\g$ denote a 3D Gaussian across the boundary.
Its scale in the long axis and the 3D center are defined as $s$ and $\mathbf{\mu}$, respectively.
We have 
\begin{gather}
     s^{\prime} = \lambda_{2D} s, \\
    \mathbf{\mu}^{\prime} = \mathbf{\mu} + \frac{1}{2}\left(s - \lambda_{2D}s\right) \mathbf{e},
\end{gather}
where $\mathbf{e}$ is the unit vector pointing from the 3D Gaussian center to the in-mask endpoint of the long axis.
The decomposed Gaussian $\g^\prime$ adopt $\mathbf{\mu}^\prime$ and $s^{\prime}$ as the new center and long-axis scale, maintaining other properties unchanged.
Another decomposed Gaussian outside of the mask is removed.


\subsection{Multiview Label Voting}
\label{sec:vote}
So far, every 3D Gaussian including the decomposed one has a list of binary labels $\mathbf{L}_i$, using the label assignment in Sec.~\ref{sec:mask_assign}.
Leveraging the assigned labels, here we adopt a simple yet effective heuristic rule to determine if a 3D Gaussian $\g_i$ belongs to the target 3D object.
In particular, we first define the confidence score $s_i$ of $g_i$ as 
\begin{equation}
    s_i = \frac{1}{N}\sum_{j =0 }^{N-1} L_{ij},
\end{equation}
where $N$ is the number of views. 
First, $s_i > 0.5$ is performed to vote out the object 3D Gaussian.
Then, to reduce the background bias due to occlusion, we adopt a threshold $\tau$ and an Object-ID $\mathbf{O}_{i}$ for each 3D Gaussian $\g_i$ is determined by:
\begin{equation}
\mathbf{O}_{i} =
    \begin{cases}
        1 & \text{if } s_i>\tau, \\
        0 & \text{if } s_i<\tau,
    \end{cases}
    \label{eq:single-obj}
\end{equation}

\subsection{Multi-object Segmentation}
There are numerous objects in the 3D scenes. Our method can also support segmenting all objects simultaneously by merely extending the binary masks to multi-label masks.
Specifically, the element $\mathbf{L}_{ij}$ in the matrix $\mathbf{L}$ is extended from binary values to multiple values, where $\mathbf{L}_{ij} \in \{0, 1, ..., C-1\}$ and $C$ is the number of objects. 
The similar multi-view label voting strategy will determine an Object-ID $\mathbf{O}_{i}$ for each 3D Gaussian $\g_i$:
\begin{equation}
   \mathbf{O}_{i} = Mode(\mathbf{L}_{ij}), j\in\{0,1,...,N-1 \},
   \label{eq:multi-obj}
\end{equation}
where $Mode(.)$ is the function of finding the element that appears the most, and $N$ is the number of views.

\section{Experimets}
\label{sec:exp}

\subsection{Datasets}

We choose different datasets to testify our method, including LLFF~\cite{llff}, Mip-NeRF 360~\cite{mip-nerf}, LERF~\cite{lerf}, and some test scenes from the 3D Gaussian Splatting~\cite{3DGS}. These datasets contain both small indoor objects and large outdoor scenes, which are very complex and challenging. For quantitative experiments, because there is no existing benchmark that can be used in 3D Gaussian space, we use the SPIn-NeRF~\cite{spin-nerf} dataset with 2D ground truth for evaluation.

\subsection{Implementation Details}

As the implementation of 3D Gaussian Splatting~\cite{3DGS} for each scene, we follow the official code with default parameters, and each scene is trained with 30000 iterations. 
For generating single object segmentation masks, we don't limit the number of clicked points on a single reference view for SAM. Binary masks of other views can be automatically generated via the proposed method. 
This is reasonable since users can refine their input prompts to help SAM generate a 2D mask as accurately as possible from the reference view. 
For scene segmentation to generate multi-label masks, we first employ SAM to produce instance masks for individual views under automatic segmentation mode. To ensure 2D mask consistency across views, we use a pre-trained zero-shot tracker~\cite{cheng2023tracking,gaussian-grouping} to propagate and associate masks.
For the selection of hyper-parameters, we only have one to control, which is the confidence score threshold $\tau$ in the Multiview Label Voting method~\ref{sec:vote}. By default, we set the value to 0.7 in all experiments. Also, in practical use, users can manually set this value according to the complexity of different scenes.
As for the number of views used in experiments, we follow SA3D~\cite{SA3D} to select all view images of each scene to finish our segmentation process. 

\subsection{Quantitative results}

\begin{table*}[!tbp]
\centering
\footnotesize
\setlength{\tabcolsep}{5.0mm}{
    \caption{Quantitative results on SPIn-NeRF dataset. \textit{Singe view} denotes projecting the 2D segmentation result to 3D simply.} \label{tab:spin-nerf}
  \begin{tabular}{c|cc|cc|cc|cc|cc}
    \toprule
    \multirow{2}{*}{Scenes} &  \multicolumn{2}{c}{Single View~\cite{SA3D}} & \multicolumn{2}{c}{MVSeg~\cite{spin-nerf}} & \multicolumn{2}{c}{SA3D~\cite{SA3D}} & \multicolumn{2}{c}{SAGA~\cite{SAGA}} & \multicolumn{2}{c}{Ours}  \\ \cline{2-11}
     & IoU & Acc & IoU & Acc & IoU & Acc & IoU & Acc & IoU & Acc\\
     \midrule
    
    Orchids & 79.4 & 96.0 & 92.7 & 98.8 & 83.6 & 96.9 & - & - & 85.4 & 97.5 \\
    Ferns & 95.2 & 99.3 & 94.3 & 99.2 & 97.1 & 99.6 & - & - & 92.0 & 98.9 \\
    Room & 73.4 & 96.5 & 95.6 & 99.4 & 88.2 & 98.3 & - & - & 86.5 & 98.1 \\
    Horns & 85.3 & 97.1 & 92.8 & 98.7 & 94.5 & 99.0 & - & - & 91.1 & 98.4  \\
    Fortress & 94.1 & 99.1 & 97.7 & 99.7 & 98.3 & 99.8 & - & - & 95.6 & 99.5 \\
    Fork & 69.4 & 98.5 & 87.9 & 99.5 & 89.4 & 99.6 & - & - & 83.4 & 99.3 \\
    Pinecone & 57.0 & 92.5 & 93.4 & 99.2 & 92.9 & 99.1 & - & - & 92.6 & 99.0 \\
    Truck & 37.9 & 77.9 & 85.2 & 95.1 & 90.8 & 96.7 & - & - & 93.0 & 97.9 \\
    Lego & 76.0 & 99.1 & 74.9 & 99.2 & 92.2 & 99.8 & - & - & 90.2 & 99.7 \\
    \midrule
    Mean & 74.1 & 95.2 & 90.5 & 98.8 & 91.9 & 98.8 & 88.0 & 98.5 & 90.0 & 98.7 \\
    \bottomrule
  \end{tabular}}
\end{table*}

\noindent \textbf{\textit{Point-Guided Segmentation:}} 
We first conduct experiments on the SPIn-NeRF~\cite{spin-nerf} dataset for quantitative analysis.
Given a set of images of the scene, we follow the process described in Section~\ref{method} to obtain the segmentation of the target object in 3D Gaussian Space.
The segmented 3D Gaussians are used to render 2D masks in other views.
Finally, we calculate the IoU and Accuracy between these rendered and the ground-truth masks. Results can be seen in Table~\ref{tab:spin-nerf}.

In the comparison, it is noteworthy that both the MVSeg~\cite{spin-nerf} and the SA3D~\cite{SA3D} require additional parameters and a computation-costly training process.
While SAGA~\cite{SAGA} shows nearly one-thousandth of the time compared with them, it needs extra training time (10 minutes per scene) to distill the knowledge embedded in the SAM decoder into the feature field.
By contrast, the ``Single view''~\cite{SA3D} refers to mapping the 2D masks to the 3D space based on the corresponding depth information, which does not need additional training process.
In this sense, our method is the same as the ``Single view'' that does not incorporate any additional training or model parameters.
Results show that our approach can achieve comparable segmentation quality to the MVSeg and SA3D methods and better results than SAGA. The time cost of ours is much less than MVSeg and SA3D. Considering the training time of SAGA, the average segmentation cost of a single object is basically the same as our method.
In some 360 outdoor scenes, such as ``Truck'', our approach even outperforms the SA3D and the MVSeg. When compared with the training-free method ``Single view'', our approach achieves a significant promotion of +15.9\% IoU and +3.5\% Acc. 
These results demonstrate our approach is very efficient in obtaining high-quality segmentation masks.

\begin{table}[!bpt]
\centering
\setlength{\tabcolsep}{2.2mm}{
\caption{Quantitative results with text prompts on four scenes of SPIn-NeRF dataset.}
\label{tab:text-prompt}
  \begin{tabular}{l|cccc}
    \toprule
     Scene & room & truck & fortress & pinecone \\
     \midrule
     Text & ``the table'' & ``the truck'' & ``the fortress'' & ``the pinecone'' \\
     \midrule
     IoU & 86.3 & 93.7 & 92.8 & 86.5 \\ 
     Acc & 97.9 & 97.8 & 98.8 & 97.0 \\     
    \bottomrule
  \end{tabular}}
\end{table}

\noindent \textbf{\textit{Text-Guided Segmentation:}}
We replace the point prompts with text prompts corresponding to the objects. Given a text input, we prompt the Grounding DINO~\cite{groundingdino} to generate target bounding boxes, which serve as input prompts for the SAM to obtain 2D segmentation masks. 
Then, following the segmentation process in Section~\ref{method}, we can transform the 2D masks into 3D Gaussian masks corresponding to the text input. 
Similar to the above evaluation process, we also calculate the IoU and Acc between 2D rendered masks and given ground-truth masks. 
Quantitative results are shown in Table~\ref{tab:text-prompt}.
For both room and truck scenes, the IoU and Acc values can achieve a similar level as using clicked points as input prompts. These results demonstrate our method can combine multi-modal prompts as input.


\subsection{Boundary Segmentation Analysis}


The main contribution of our work is to address the boundary issues of segmentation in 3DGS, as mentioned in Sec.~\ref{sec:intro} and Fig.~\ref{fig:intro}.
To validate the effectiveness of our work, we evaluate metrics, including IoU, AP, and F1-score, on region nearby boundaries.
Specifically, we extract parts of 3-pixel width alongside boundaries of ground-truth masks and compare results with previous work SAGA~\cite{SAGA}. Experiments are conducted on Scene \textit{pinecone} and \textit{fortress} from the LLFF~\cite{llff} dataset. Better results of ours are indicated in Tab.~\ref{tab:boundary-mask}. Benefiting from our proposed Gaussian Decomposition for ambiguous geometry around boundaries, we can achieve more complete and accurate boundary segmentation results.

\begin{figure*}[!tbp]
    \centering
    \includegraphics[width=0.99\linewidth]{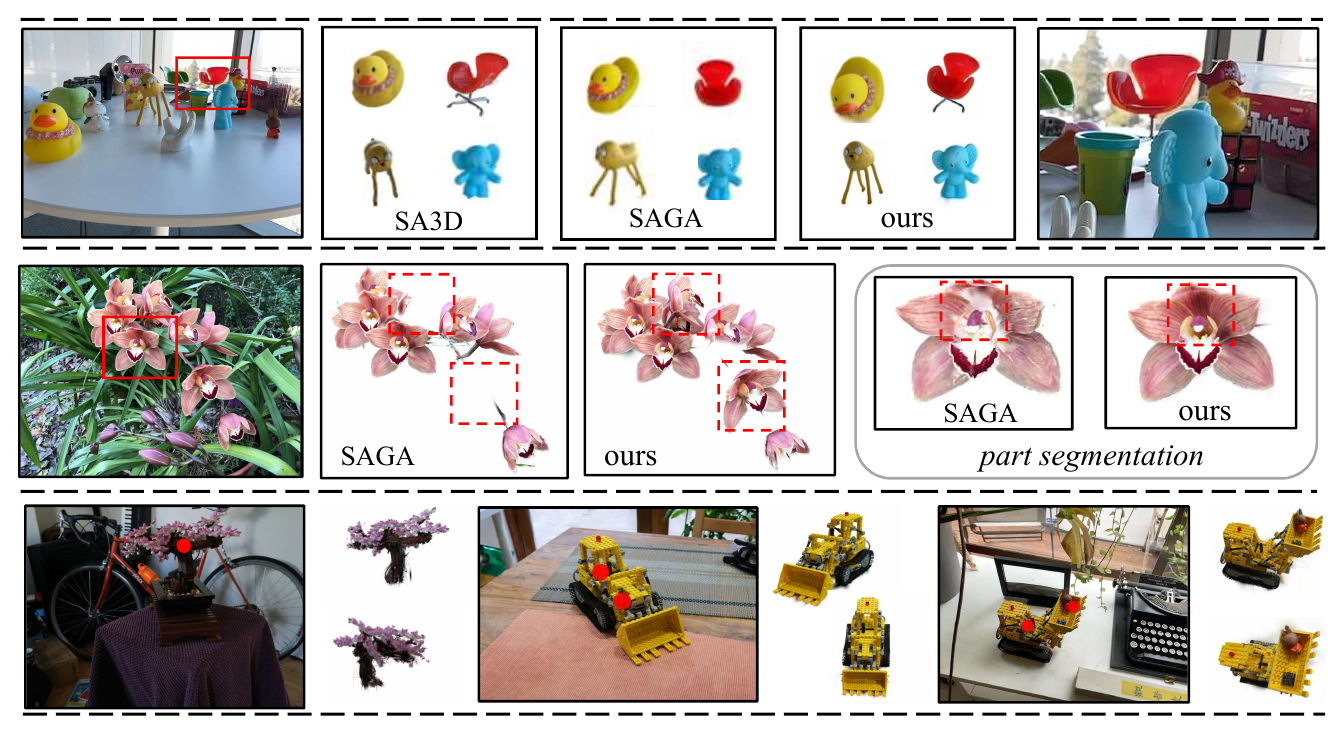}
    \caption{Qualitative results compared with SA3D~\cite{SA3D} and SAGA~\cite{SAGA} in different scenes (LERF-figurines~\cite{lerf}, SPIn-NeRF-Orchids~\cite{spin-nerf}, LERF-dozer-nerfgun-waldo~\cite{lerf}). We enlarge the boxed area on the right for a better visualization.}
    \label{fig:overall}
\end{figure*}

\begin{table}[!bpt]
\centering
\setlength{\tabcolsep}{2.5mm}{
\caption{Evaluation on boundary masks. We extract parts of \textbf{3-pixel}
width alongside boundaries of ground truth and segmentation masks and evaluate on \textbf{IoU, AP, and F1-score} metrics.}
\label{tab:boundary-mask}
  \begin{tabular}{c|cccccc}
    \toprule
      \multirow{2}{*}{Scene} & \multicolumn{2}{c}{IoU} & \multicolumn{2}{c}{AP} & \multicolumn{2}{c}{F1-score}   \\ \cline{2-7}
       & SAGA~\cite{SAGA} & Ours & SAGA & Ours & SAGA & Ours \\  \midrule
       pinecone & 13.8 & \textbf{21.8} & 15.7 & \textbf{23.5} & 24.0 & \textbf{35.6} \\
       fortress & 16.7 & \textbf{23.3} & 31.8 & \textbf{60.2} & 28.5 & \textbf{37.2}\\
    \bottomrule
  \end{tabular}}
\end{table}

\subsection{Qualitative results}

We conduct four kinds of tasks to demonstrate the potential application and qualitative performance of our approach, including point-guided segmentation, text-guided segmentation, scene editing, and collision detection.

\begin{figure}[!tbp]
    \centering
    \includegraphics[width=0.99\linewidth]
    {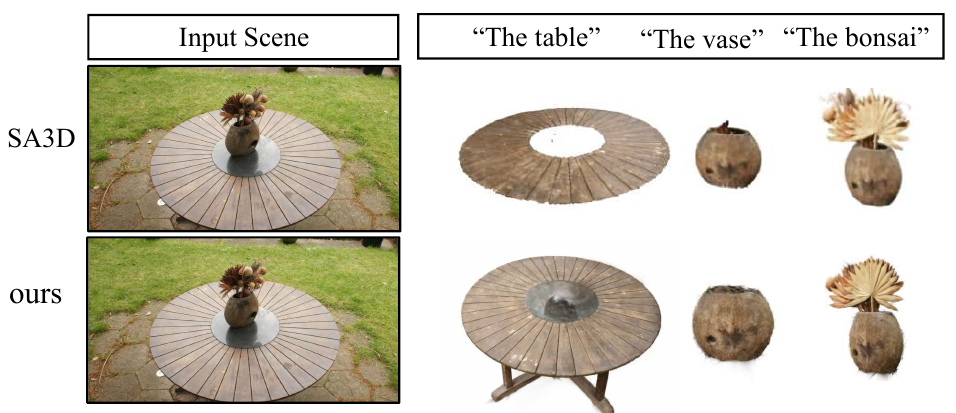}
    \caption{3D segmentation with text prompts in Mip-NeRF360-garden~\cite{mip-nerf}.}
    \label{fig:text}
\end{figure}

\noindent \textbf{\textit{Point-Guided Segmentation:}}
We first conduct visualization experiments on 3D object segmentation guided by one or a set of point prompts clicked on the first-view rendered image. The results are shown in Fig.~\ref{fig:overall}.
The first row shows the segmentation results of SA3D, SAGA and ours on LERF-figurines scene. The second row compares with SAGA, which distills the knowledge embedded in the SAM decoder into the feature field. Though the boundary issues in 3D-GS can be alleviated, incomplete segmentation occurs even after post-processing. In contrast, ours can solve the boundary problems while achieving more complete segmentation. More segmentation results are shown in the third row.


\noindent \textbf{\textit{Text-Guided Segmentation:}}
We further conduct 3D object segmentation experiments guided by text prompts. 
We follow the same process as in Table~\ref{tab:text-prompt} experiments to conduct visualization experiments.
In order to compare our method with SA3D under the same setting, we choose the same garden scene from Mip-NeRF 360 dataset~\cite{mip-nerf}, using three text prompts including ``The table'', ``The vase'' and ``The bonsai''. 
Results in Fig.~\ref{fig:text} show that our approach can achieve accurate segmentation results by simply providing object names, demonstrating our method's potential in combining with language models. 
Compared with the SA3D in the ``The table'' case, our approach can segment the complete object with desired table legs, while SA3D only gives the tabletop.

\begin{figure}[!tbp]
    \centering
    \includegraphics[width=0.99\linewidth]
    {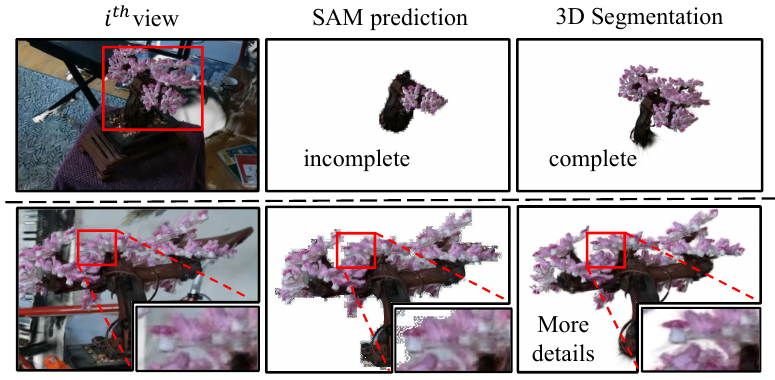}
    \caption{Comparison between 2D and 3D Segmentation. Noisy predictions exist in SAM's segmentation, such as incomplete objects and background regions. These issues are solved in our 3D segmentation.}
    \label{fig:2d_3d}
\end{figure}

\noindent \textbf{\textit{Comparison between 2D and 3D Segmentation:}}
Our method improves the segmentation quality of SAM.
Due to the sensitivity of SAM to the scene viewpoint (e.g. illumination), the 2D segmentation results might be incomplete under some views, but our 3D segmentation can still guarantee complete segmentation results via a multi-view ensemble (Fig.~\ref{fig:2d_3d}, $1^{st}$ Row).
Furthermore, we can obtain detailed 3D segmentation results under noisy predictions from SAM. 
As shown in Fig.~\ref{fig:2d_3d}, $2^{nd}$ Row, the multi-view voting ensures that object Gaussians are retained correctly and background Gaussians obtained from the noisy boundaries of SAM predictions are removed.

\begin{figure*}[!tbp]
    \centering
    \includegraphics[width=0.98\linewidth]
    {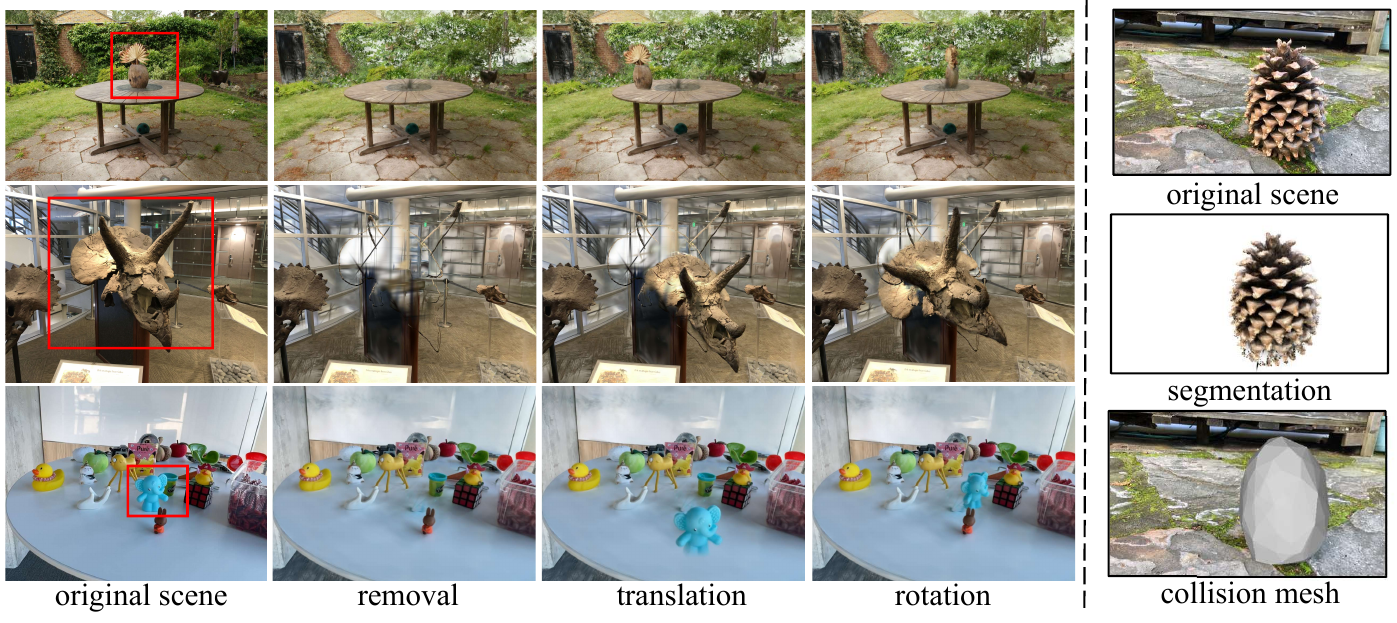}
    \caption{Visualization examples of scene editing after the object segmentation. We offer three scene editing examples: removal, translation, and rotation. We further provide collision mesh computed after segmentation, as shown in the last column.}
    \label{fig:applications}
\end{figure*}

\noindent \textbf{\textit{Scene Editing:}}
Scene editing is a basic application for 3D reconstructed scenes. However, it's quite difficult to do this without being able to locate the specific objects.
This task demonstrates the ability of our approach to help edit the 3D scenes. Specifically, after segmenting the objects in the 3D Gaussian space, we can manipulate the objects by removing, translating, and rotating them. 
Thanks to the simplicity of the explicit Gaussian representations, without bells and whistles, our approach obtains satisfactory scene editing results, as shown in Fig.~\ref{fig:applications}.
The instances in 1st column with red bounding boxes are objects to be segmented. It can be seen that with objects segmented in 3D Gaussians, they can be translated and rotated in any direction in the scene. After the removal of segmented objects, original scenes can still keep intact. 


\noindent \textbf{\textit{Collision Detection:}}
Collision detection is an indispensable component in practical 3D applications, such as games, movies, and simulators. In this task, we demonstrate our approach can directly help in revealing the collision body of target objects in the 3D Gaussian space. 
We choose two scenes from SPIn-NeRF dataset. Following the process of our segmentation method, we can obtain the corresponding segmented 3D Gaussian points in the scene.
To this end, we use the Quickhull~\cite{quickhull} algorithm to build the collision mesh upon our segmented objects. The results in the last column of Fig.~\ref{fig:applications} show that our segmented objects can successfully derive correct convex hulls for downstream applications.

\subsection{Time Cost Analysis}


We conduct the time cost analysis compared with previous 3DGS segmentation methods, namely SAGA~\cite{SAGA} and Gaussian Grouping~\cite{gaussian-grouping}. 
We select the \textit{figurines} scene from the LERF dataset~\cite{lerf}, which includes 20+ objects, using a single NVIDIA A100 GPU for fair evaluation.
All methods utilize SAM to extract image features and then perform mask generation. Thus, they share a similar time cost, roughly 2 minutes to process 300 images (\textit{SAM Extraction} in Tab.~\ref{tab:time-cost}). 
This procedure can be regarded as a pre-processing and can be accelerated by image-batching or deploying Efficient-SAM.
For extra processing time, both baseline methods require gradient descent optimization over 30,000 iterations to distill SAM features~\cite{SAGA} or ID embedding~\cite{gaussian-grouping} into 3D space linked with each 3D Gaussian, resulting in considerable additional training time for re-optimizing 3D scenes. On the contrary, our method only needs to perform multi-view and multi-label assignments, the time cost of which is merely 12 seconds.
As for segmenting a single object, our method merely requires an efficient voting process, taking just 0.4 milliseconds, which is much faster than other methods as they both demand network forwarding.

\begin{table}[!bpt]
\centering
\setlength{\tabcolsep}{1.5mm}{
\caption{Time cost comparisions on \textit{figurines} scene with 20+ objects to be segmented simultaneously.}
\label{tab:time-cost}
  \begin{tabular}{c|cccc}
    \toprule
     \multirow{2}{*}{Time} & SAM  & Extra  & Optimization  & Forward  \\
                           & Extraction & Processing & Steps & Segmentation \\
     \midrule
     SAGA~\cite{SAGA} & \multirow{3}{*}{2 min} & 10 min & 30000 & 0.5s \\ 
     Ga-Grouping~\cite{gaussian-grouping} &  & 9 min & 30000 & 0.3s \\ 
     Ours & & 12s & 1 & 0.4ms \\
    \bottomrule
  \end{tabular}}
\end{table}

\begin{table}[!bpt]
\centering
\setlength{\tabcolsep}{3.2mm}{
    \caption{Ablation on Gaussian decomposition (GD) on SPIn-NeRF dataset. The 2nd and 3rd columns compare the results before and after using the Gaussian decomposition (GD). The last column represents the results of directly removing the gaussians across mask boundaries.} \label{tab:gaussian-decomposition}
  \begin{tabular}{l|cccccc}
    \toprule
     \multirow{2}{*}{Scene} & \multicolumn{2}{c}{w/ GD} & \multicolumn{2}{c}{w/o GD} & \multicolumn{2}{c}{Delete}\\
     \cline{2-7}
     & IoU & Acc & IoU & Acc & IoU & Acc \\ 
     \midrule
     Orchids & 85.4 & 97.5 & 82.2 & 96.8 & 78.2 & 95.4 \\
     Ferns & 92.0 & 98.9 & 89.2 & 98.4 & 89.8 & 98.5 \\
     Room & 86.5 & 98.1 & 81.3 & 97.2 & 85.4 & 97.9  \\
     Horns & 91.1 & 98.4 & 83.2 & 96.5 & 84.8 & 97.6 \\
     Fortress & 96.5 & 99.4 & 88.5 & 98.1 & 82.3 & 96.7 \\
     Fork & 83.4 & 99.3 & 81.8 & 99.2 & 79.9 & 99.1 \\
     Pinecone & 92.6 & 98.9 & 91.6 & 98.9 & 82.9 & 97.5  \\
     Truck & 93.0 & 97.9 & 93.4 & 97.8 & 91.4 & 96.8 \\
     Lego & 90.2 & 99.7 & 88.4 & 99.6 & 82.9 & 99.4 \\ 
     \midrule
     mean & \textbf{90.0} & \textbf{98.7} & 86.6 & 98.1 & 84.2 & 97.7 \\
    \bottomrule
  \end{tabular}}
\end{table}

\subsection{Ablation Study}

\noindent \textbf{\textit{Gaussian Decomposition:}}
Gaussian Decomposition is proposed to address the issue of roughness boundaries of 3D segmented objects, which results from the non-negligible spatial sizes of 3D Gaussian located at the boundary. We conduct both quantitative and qualitative experiments to verify the effectiveness of our approach.
For quantitative ablation experiments, Table~\ref{tab:gaussian-decomposition} shows the results.
We compare our proposed Gaussian Decomposition scheme with two other processing methods, one for segmentation without any special handling (the 3rd column), and the other for directly removing such Gaussians (the last column). The experiments are conducted on SPIn-NeRF dataset, following the same evaluation process as Table~\ref{tab:spin-nerf}.
Compared with the others, our proposed approach outperforms them on all scenes with +3.4\% and +5.8\% IoU, respectively, demonstrating the effectiveness of Gaussian Decomposition. Besides, it can seen that directly removing these gaussians cannot decrease the roughness of mask boundaries, leading to even 2.4\% lower IoU values.

\begin{figure}[!tbp]
    \centering
    \includegraphics[width=0.99\linewidth]{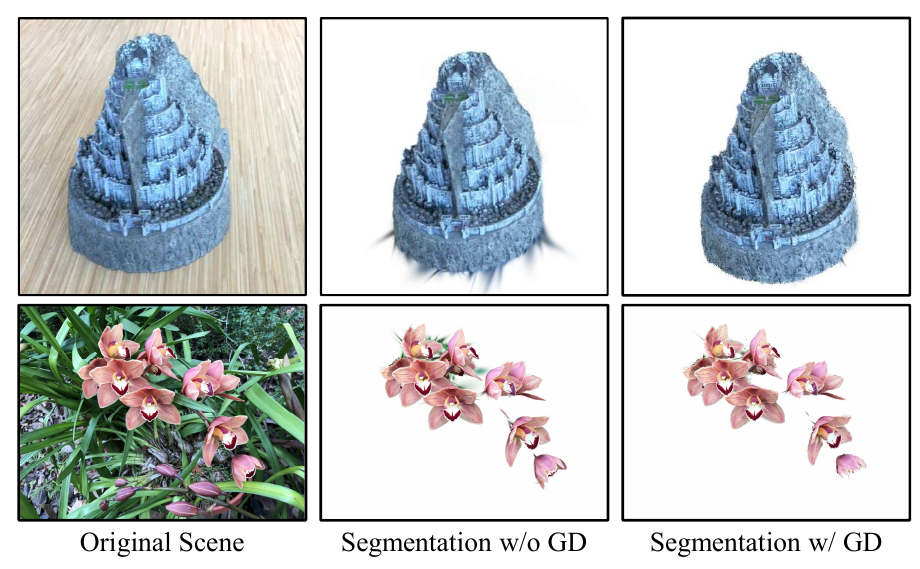}
    \caption{Ablation results on Gaussian Decomposition (GD) on the LERF-pinecone.}
    \label{fig:gaussian-decomposition}
     \vspace{-0.5cm}
\end{figure}

Fig.~\ref{fig:gaussian-decomposition} shows the visualization results for comparison. We select two representative scenes in which large-scale Gaussian points exist at the junction/contact (can be seen in the 2nd column). This issue is alleviated after utilizing the proposed Gaussian Decomposition strategy. Though this idea is simple, the improvement is obvious.

\begin{table}[!bpt]
\centering
\setlength{\tabcolsep}{2.5mm}{
    \caption{Ablation on different number of views for 3D segmentation. Numbers in parentheses represent the used view percentage of the total training view.}
    \label{tab:number-views}
  \begin{tabular}{l|cccc}
    \toprule
     Number of views & 5(10\%) & 10(20\%) & 21(50\%) & 42(100\%) \\
     IoU on Fortress & 91.16 & 92.11 & 93.82 & 95.6  \\ 
     \midrule
    Number of views & 25(10\%) & 50(20\%) & 125(50\%) & 251(100\%) \\
    IoU on truck & 90.27 & 90.97 & 92.11 & 93.0  \\ 
    \bottomrule
  \end{tabular}}
\end{table}


\begin{table}[!tbp]
\centering
\setlength{\tabcolsep}{5.0mm}{
\caption{Ablation on confidence score threshold $\tau$ on the \textit{truck} scene.}
\label{tab:confidence-score}
  \begin{tabular}{l|cccc}
    \toprule
     Scene / $\tau$ & 0.6 & 0.65 & 0.7 & 0.8  \\ 
     \midrule
     pinecone & 90.8 & 92.0 & \textbf{92.6} & 91.4 \\
     fortress & 94.0 & 95.4 & 95.6 & \textbf{96.2} \\
     lego & 89.7 & 89.8 & \textbf{90.2} & \textbf{90.2} \\
    \bottomrule
  \end{tabular}}
\end{table}




\begin{figure*}[!tbp]
    \centering
    \includegraphics[width=0.98\linewidth]
    {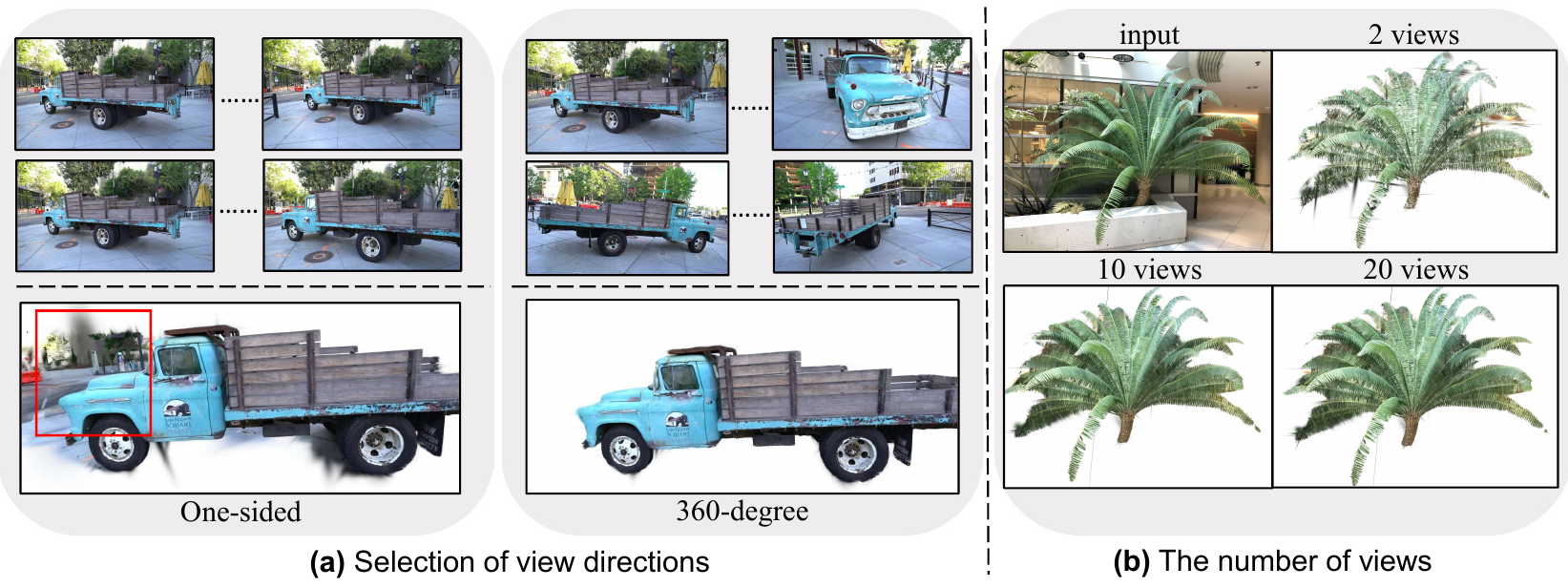}
    \caption{Ablation on view selection, including view directions and the number of views.}
    \label{fig:selection_views}
\end{figure*}

\noindent \textbf{\textit{Selection of Views:}}
In this section, we study the influence of selected 2D views on 3D Gaussian segmentation.
Fig.~\ref{fig:selection_views} (b) demonstrates the segmentation quality using different numbers of views. It is noteworthy that the leaves in the ``fern'' scene are very tiny and challenging. Even so, with only two sparse views, our approach can achieve decent results, and the segmentation quality quickly improves when increasing the number of views. 
Table~\ref{tab:number-views} also draws a similar conclusion: as the number of views increases, the IoU values will also increase. It is worth noting that even with sparse views (10\% percentage), we can obtain relatively decent results for both two types of scenes, which also demonstrates the robustness and effectiveness of our approach. 
In practical applications, using sparse views (below 10\% of total views) will greatly improve efficiency, although under 100\% views, our method can still be completed within one minute.

Fig.~\ref{fig:selection_views} (a) study the influence of view directions on final segmentation results.
We choose the truck with 360-degree views.
The upper row shows the result of using roughly a single direction. In this case, the background is mistakenly incorporated into the segmentation. By contrast, with the same number of views but variant directions, the approach achieves high-quality clean segmentation results. This result reveals the importance of choosing non-monotone view directions in practice usage with 360 scenes.



\noindent \textbf{\textit{Hyper-Parameters:}}
In the process of our method, we set a hyper-parameter $\tau$ to control the threshold of label confidence score in Sec.~\ref{sec:vote}. To evaluate the robustness of the selection of the hyper-parameter under different scenes, we set different values to conduct experiments on Scene \textit{pinecone}, \textit{fortress}, and \textit{lego} from LLFF~\cite{llff} dataset.
As indicated in Tab.~\ref{tab:confidence-score}, our method is not sensitive to the threshold selection. Robust results are observed in multiple scenes when setting the threshold between 0.6 and 0.8. Values within this range can obtain fine results. In all of our experiments, the value of the hyper-parameter $\tau$ is set to 0.7 by default.

\section{Discussions and Limitations}
Through experiments, we found that our method does not perform well in objects where 3D Gaussians are very sparse, such as the LLFF-room~\cite{llff} scene.
The Gaussians of the table are notably sparse; even worse, the Gaussians representing the table surface are across different objects. Though our Gaussian Decomposition can somewhat remedy the boundary issue, it still suffers from extremely sparse points and leaves holes in 3D segmentation. 
We believe this limitation can be alleviated by future research in a more structured 3D-GS representation, yielding more accurate results.

\section{Conclusion}

We address the issue that rough/incomplete boundaries in segmenting 3D-GS and propose a novel Boundary-enhanced segmentation pipeline.
Given input prompts on the first rendering view, our approach automatically generates multi-view masks and achieves consistent 3D segmentation via the proposed assignment strategy.
Our Gaussian Decomposition module can effectively mitigate the boundary roughness issue of segmented objects resulting from the inherent geometry structure in 3D-GS.
Extensive segmentation experiments show that our method can effectively obtain high-quality 3D object segmentation without boundary issues, and different scene-editing tasks demonstrate that our method can be easily applied to downstream applications. 
Overall, we hope our work can inspire more future work in the area of 3D Gaussian representation.

{

\bibliographystyle{plain}
\bibliography{main}

\begin{thebibliography}{100}

\bibitem{bae2024per}
Jeongmin Bae, Seoha Kim, Youngsik Yun, Hahyun Lee, Gun Bang, and Youngjung Uh.
\newblock Per-gaussian embedding-based deformation for deformable 3d gaussian splatting.
\newblock {\em arXiv preprint arXiv:2404.03613}, 2024.

\bibitem{quickhull}
C~Bradford Barber, David~P Dobkin, and Hannu Huhdanpaa.
\newblock The quickhull algorithm for convex hulls.
\newblock {\em ACM Transactions on Mathematical Software (TOMS)}, 22(4):469--483, 1996.

\bibitem{mip-nerf}
Jonathan~T Barron, Ben Mildenhall, Dor Verbin, Pratul~P Srinivasan, and Peter Hedman.
\newblock Mip-nerf 360: Unbounded anti-aliased neural radiance fields.
\newblock In {\em Proceedings of the IEEE/CVF Conference on Computer Vision and Pattern Recognition}, pages 5470--5479, 2022.

\bibitem{bolya2019yolact}
Daniel Bolya, Chong Zhou, Fanyi Xiao, and Yong~Jae Lee.
\newblock Yolact: Real-time instance segmentation.
\newblock In {\em Proceedings of the IEEE/CVF international conference on computer vision}, pages 9157--9166, 2019.

\bibitem{bu2021gaia}
Xingyuan Bu, Junran Peng, Junjie Yan, Tieniu Tan, and Zhaoxiang Zhang.
\newblock Gaia: A transfer learning system of object detection that fits your needs.
\newblock In {\em Proceedings of the IEEE/CVF Conference on Computer Vision and Pattern Recognition}, pages 274--283, 2021.

\bibitem{SAGA}
Jiazhong Cen, Jiemin Fang, Chen Yang, Lingxi Xie, Xiaopeng Zhang, Wei Shen, and Qi~Tian.
\newblock Segment any 3d gaussians.
\newblock {\em CoRR}, abs/2312.00860, 2023.

\bibitem{SA3D}
Jiazhong Cen, Zanwei Zhou, Jiemin Fang, Wei Shen, Lingxi Xie, Xiaopeng Zhang, and Qi~Tian.
\newblock Segment anything in 3d with nerfs.
\newblock {\em arXiv preprint arXiv:2304.12308}, 2023.

\bibitem{chang2022data}
Qing Chang, Junran Peng, Lingxi Xie, Jiajun Sun, Haoran Yin, Qi~Tian, and Zhaoxiang Zhang.
\newblock Data: Domain-aware and task-aware self-supervised learning.
\newblock In {\em Proceedings of the IEEE/CVF Conference on Computer Vision and Pattern Recognition}, pages 9841--9850, 2022.

\bibitem{charatan2024pixelsplat}
David Charatan, Sizhe~Lester Li, Andrea Tagliasacchi, and Vincent Sitzmann.
\newblock pixelsplat: 3d gaussian splats from image pairs for scalable generalizable 3d reconstruction.
\newblock In {\em Proceedings of the IEEE/CVF Conference on Computer Vision and Pattern Recognition}, pages 19457--19467, 2024.

\bibitem{TensorRF}
Anpei Chen, Zexiang Xu, Andreas Geiger, Jingyi Yu, and Hao Su.
\newblock Tensorf: Tensorial radiance fields.
\newblock In {\em Proc. ECCV}, pages 333--350. Springer, 2022.

\bibitem{chen2024rsprompter}
Keyan Chen, Chenyang Liu, Hao Chen, Haotian Zhang, Wenyuan Li, Zhengxia Zou, and Zhenwei Shi.
\newblock Rsprompter: Learning to prompt for remote sensing instance segmentation based on visual foundation model.
\newblock {\em IEEE TGRS}, 2024.

\bibitem{chen2014semantic}
Liang-Chieh Chen, George Papandreou, Iasonas Kokkinos, Kevin Murphy, and Alan~L Yuille.
\newblock Semantic image segmentation with deep convolutional nets and fully connected crfs.
\newblock {\em arXiv preprint arXiv:1412.7062}, 2014.

\bibitem{chen2017deeplab}
Liang-Chieh Chen, George Papandreou, Iasonas Kokkinos, Kevin Murphy, and Alan~L Yuille.
\newblock Deeplab: Semantic image segmentation with deep convolutional nets, atrous convolution, and fully connected crfs.
\newblock {\em IEEE transactions on pattern analysis and machine intelligence}, 40(4):834--848, 2017.

\bibitem{chen2017rethinking}
Liang-Chieh Chen, George Papandreou, Florian Schroff, and Hartwig Adam.
\newblock Rethinking atrous convolution for semantic image segmentation.
\newblock {\em arXiv preprint arXiv:1706.05587}, 2017.

\bibitem{chen2018encoder}
Liang-Chieh Chen, Yukun Zhu, George Papandreou, Florian Schroff, and Hartwig Adam.
\newblock Encoder-decoder with atrous separable convolution for semantic image segmentation.
\newblock In {\em Proceedings of the European conference on computer vision (ECCV)}, pages 801--818, 2018.

\bibitem{chen2024text}
Zilong Chen, Feng Wang, Yikai Wang, and Huaping Liu.
\newblock Text-to-3d using gaussian splatting.
\newblock In {\em Proceedings of the IEEE/CVF Conference on Computer Vision and Pattern Recognition}, pages 21401--21412, 2024.

\bibitem{cheng2022maskedattention}
Bowen Cheng, Ishan Misra, Alexander~G Schwing, Alexander Kirillov, and Rohit Girdhar.
\newblock Masked-attention mask transformer for universal image segmentation.
\newblock In {\em Proceedings of the IEEE/CVF Conference on Computer Vision and Pattern Recognition}, pages 1290--1299, 2022.

\bibitem{cheng2021maskformer}
Bowen Cheng, Alexander~G. Schwing, and Alexander Kirillov.
\newblock Per-pixel classification is not all you need for semantic segmentation.
\newblock In {\em NeurIPS}, 2021.

\bibitem{cheng2023tracking}
Ho~Kei Cheng, Seoung~Wug Oh, Brian Price, Alexander Schwing, and Joon-Young Lee.
\newblock Tracking anything with decoupled video segmentation.
\newblock In {\em Proceedings of the IEEE/CVF International Conference on Computer Vision}, pages 1316--1326, 2023.

\bibitem{surfels}
Pinxuan Dai, Jiamin Xu, Wenxiang Xie, Xinguo Liu, Huamin Wang, and Weiwei Xu.
\newblock High-quality surface reconstruction using gaussian surfels.
\newblock In {\em ACM SIGGRAPH 2024 Conference Papers}, pages 1--11, 2024.

\bibitem{deng2023segment}
Ruining Deng, Can Cui, Quan Liu, Tianyuan Yao, Lucas~W Remedios, Shunxing Bao, Bennett~A Landman, Lee~E Wheless, Lori~A Coburn, Keith~T Wilson, et~al.
\newblock Segment anything model (sam) for digital pathology: Assess zero-shot segmentation on whole slide imaging.
\newblock {\em arXiv preprint arXiv:2304.04155}, 2023.

\bibitem{FSD}
Lue Fan, Feng Wang, Naiyan Wang, and ZHAO-XIANG ZHANG.
\newblock Fully sparse 3d object detection.
\newblock {\em Advances in Neural Information Processing Systems}, 35:351--363, 2022.

\bibitem{RangeDet}
Lue Fan, Xuan Xiong, Feng Wang, Naiyan Wang, and Zhaoxiang Zhang.
\newblock Rangedet: In defense of range view for lidar-based 3d object detection.
\newblock In {\em Proc. ICCV}, pages 2918--2927, 2021.

\bibitem{plenoxel}
Sara Fridovich-Keil, Alex Yu, Matthew Tancik, Qinhong Chen, Benjamin Recht, and Angjoo Kanazawa.
\newblock Plenoxels: Radiance fields without neural networks.
\newblock In {\em Proceedings of the IEEE/CVF conference on computer vision and pattern recognition}, pages 5501--5510, 2022.

\bibitem{fastnerf}
Stephan~J Garbin, Marek Kowalski, Matthew Johnson, Jamie Shotton, and Julien Valentin.
\newblock Fastnerf: High-fidelity neural rendering at 200fps.
\newblock In {\em Proceedings of the IEEE/CVF international conference on computer vision}, pages 14346--14355, 2021.

\bibitem{isrf}
Rahul Goel, Dhawal Sirikonda, Saurabh Saini, and PJ~Narayanan.
\newblock Interactive segmentation of radiance fields.
\newblock In {\em Proceedings of the IEEE/CVF Conference on Computer Vision and Pattern Recognition}, pages 4201--4211, 2023.

\bibitem{guedon2024sugar}
Antoine Gu{\'e}don and Vincent Lepetit.
\newblock Sugar: Surface-aligned gaussian splatting for efficient 3d mesh reconstruction and high-quality mesh rendering.
\newblock In {\em Proceedings of the IEEE/CVF Conference on Computer Vision and Pattern Recognition}, pages 5354--5363, 2024.

\bibitem{hafiz2020survey}
Abdul~Mueed Hafiz and Ghulam~Mohiuddin Bhat.
\newblock A survey on instance segmentation: state of the art.
\newblock {\em International journal of multimedia information retrieval}, 9(3):171--189, 2020.

\bibitem{hu2021towards}
Qingyong Hu, Bo~Yang, Sheikh Khalid, Wen Xiao, Niki Trigoni, and Andrew Markham.
\newblock Towards semantic segmentation of urban-scale 3d point clouds: A dataset, benchmarks and challenges.
\newblock In {\em Proc. CVPR}, pages 4977--4987, 2021.

\bibitem{huang20242d}
Binbin Huang, Zehao Yu, Anpei Chen, Andreas Geiger, and Shenghua Gao.
\newblock 2d gaussian splatting for geometrically accurate radiance fields.
\newblock In {\em ACM SIGGRAPH 2024 Conference Papers}, pages 1--11, 2024.

\bibitem{huang2024sc}
Yi-Hua Huang, Yang-Tian Sun, Ziyi Yang, Xiaoyang Lyu, Yan-Pei Cao, and Xiaojuan Qi.
\newblock Sc-gs: Sparse-controlled gaussian splatting for editable dynamic scenes.
\newblock In {\em Proceedings of the IEEE/CVF Conference on Computer Vision and Pattern Recognition}, pages 4220--4230, 2024.

\bibitem{ConcealedSAM}
Ge-Peng Ji, Deng-Ping Fan, Peng Xu, Bowen Zhou, Ming-Ming Cheng, and Luc Van~Gool.
\newblock Sam struggles in concealed scenes--empirical study on" segment anything".
\newblock {\em SCIS}, 2023.

\bibitem{TensorIR}
Haian Jin, Isabella Liu, Peijia Xu, Xiaoshuai Zhang, Songfang Han, Sai Bi, Xiaowei Zhou, Zexiang Xu, and Hao Su.
\newblock Tensoir: Tensorial inverse rendering.
\newblock In {\em Proc. CVPR}, pages 165--174, 2023.

\bibitem{hqsam}
Lei Ke, Mingqiao Ye, Martin Danelljan, Yu-Wing Tai, Chi-Keung Tang, Fisher Yu, et~al.
\newblock Segment anything in high quality.
\newblock In {\em NeurIPS}, 2023.

\bibitem{3DGS}
Bernhard Kerbl, Georgios Kopanas, Thomas Leimk{\"u}hler, and George Drettakis.
\newblock 3d gaussian splatting for real-time radiance field rendering.
\newblock {\em ACM Transactions on Graphics}, 42(4):1--14, 2023.

\bibitem{lerf}
Justin Kerr, Chung~Min Kim, Ken Goldberg, Angjoo Kanazawa, and Matthew Tancik.
\newblock Lerf: Language embedded radiance fields.
\newblock In {\em Proceedings of the IEEE/CVF International Conference on Computer Vision}, pages 19729--19739, 2023.

\bibitem{Kirillov_2019_CVPR}
Alexander Kirillov, Kaiming He, Ross Girshick, Carsten Rother, and Piotr Dollar.
\newblock Panoptic segmentation.
\newblock In {\em Proceedings of the IEEE/CVF Conference on Computer Vision and Pattern Recognition (CVPR)}, June 2019.

\bibitem{SAM}
Alexander Kirillov, Eric Mintun, Nikhila Ravi, Hanzi Mao, Chloe Rolland, Laura Gustafson, Tete Xiao, Spencer Whitehead, Alexander~C Berg, Wan-Yen Lo, et~al.
\newblock Segment anything.
\newblock {\em arXiv preprint arXiv:2304.02643}, 2023.

\bibitem{lai2022stratified}
Xin Lai, Jianhui Liu, Li~Jiang, Liwei Wang, Hengshuang Zhao, Shu Liu, Xiaojuan Qi, and Jiaya Jia.
\newblock Stratified transformer for 3d point cloud segmentation.
\newblock In {\em Proc. CVPR}, pages 8500--8509, 2022.

\bibitem{2d-guided}
Kun Lan, Haoran Li, Haolin Shi, Wenjun Wu, Yong Liao, Lin Wang, and Pengyuan Zhou.
\newblock 2d-guided 3d gaussian segmentation.
\newblock {\em arXiv preprint arXiv:2312.16047}, 2023.

\bibitem{li2024spacetime}
Zhan Li, Zhang Chen, Zhong Li, and Yi~Xu.
\newblock Spacetime gaussian feature splatting for real-time dynamic view synthesis.
\newblock In {\em Proceedings of the IEEE/CVF Conference on Computer Vision and Pattern Recognition}, pages 8508--8520, 2024.

\bibitem{BEAFormer}
Zhiqi Li, Wenhai Wang, Hongyang Li, Enze Xie, Chonghao Sima, Tong Lu, Yu~Qiao, and Jifeng Dai.
\newblock Bevformer: Learning bird’s-eye-view representation from multi-camera images via spatiotemporal transformers.
\newblock In {\em Proc. ECCV}, pages 1--18. Springer, 2022.

\bibitem{liang2024luciddreamer}
Yixun Liang, Xin Yang, Jiantao Lin, Haodong Li, Xiaogang Xu, and Yingcong Chen.
\newblock Luciddreamer: Towards high-fidelity text-to-3d generation via interval score matching.
\newblock In {\em Proceedings of the IEEE/CVF Conference on Computer Vision and Pattern Recognition}, pages 6517--6526, 2024.

\bibitem{liang2024gs}
Zhihao Liang, Qi~Zhang, Ying Feng, Ying Shan, and Kui Jia.
\newblock Gs-ir: 3d gaussian splatting for inverse rendering.
\newblock In {\em Proceedings of the IEEE/CVF Conference on Computer Vision and Pattern Recognition}, pages 21644--21653, 2024.

\bibitem{lin2024gaussian}
Youtian Lin, Zuozhuo Dai, Siyu Zhu, and Yao Yao.
\newblock Gaussian-flow: 4d reconstruction with dynamic 3d gaussian particle.
\newblock In {\em Proceedings of the IEEE/CVF Conference on Computer Vision and Pattern Recognition}, pages 21136--21145, 2024.

\bibitem{ling2024align}
Huan Ling, Seung~Wook Kim, Antonio Torralba, Sanja Fidler, and Karsten Kreis.
\newblock Align your gaussians: Text-to-4d with dynamic 3d gaussians and composed diffusion models.
\newblock In {\em Proceedings of the IEEE/CVF Conference on Computer Vision and Pattern Recognition}, pages 8576--8588, 2024.

\bibitem{groundingdino}
Shilong Liu, Zhaoyang Zeng, Tianhe Ren, Feng Li, Hao Zhang, Jie Yang, Chunyuan Li, Jianwei Yang, Hang Su, Jun Zhu, et~al.
\newblock Grounding dino: Marrying dino with grounded pre-training for open-set object detection.
\newblock {\em arXiv preprint arXiv:2303.05499}, 2023.

\bibitem{liu2018path}
Shu Liu, Lu~Qi, Haifang Qin, Jianping Shi, and Jiaya Jia.
\newblock Path aggregation network for instance segmentation.
\newblock In {\em Proceedings of the IEEE conference on computer vision and pattern recognition}, pages 8759--8768, 2018.

\bibitem{liu2025citygaussian}
Yang Liu, Chuanchen Luo, Lue Fan, Naiyan Wang, Junran Peng, and Zhaoxiang Zhang.
\newblock Citygaussian: Real-time high-quality large-scale scene rendering with gaussians.
\newblock In {\em European Conference on Computer Vision}, pages 265--282. Springer, 2025.

\bibitem{liu2024citygaussianv2}
Yang Liu, Chuanchen Luo, Zhongkai Mao, Junran Peng, and Zhaoxiang Zhang.
\newblock Citygaussianv2: Efficient and geometrically accurate reconstruction for large-scale scenes.
\newblock {\em arXiv preprint arXiv:2411.00771}, 2024.

\bibitem{lu20243d}
Zhicheng Lu, Xiang Guo, Le~Hui, Tianrui Chen, Min Yang, Xiao Tang, Feng Zhu, and Yuchao Dai.
\newblock 3d geometry-aware deformable gaussian splatting for dynamic view synthesis.
\newblock In {\em Proceedings of the IEEE/CVF Conference on Computer Vision and Pattern Recognition}, pages 8900--8910, 2024.

\bibitem{luiten2023dynamic}
Jonathon Luiten, Georgios Kopanas, Bastian Leibe, and Deva Ramanan.
\newblock Dynamic 3d gaussians: Tracking by persistent dynamic view synthesis.
\newblock {\em arXiv preprint arXiv:2308.09713}, 2023.

\bibitem{medicalSAM}
Jun Ma, Yuting He, Feifei Li, Lin Han, Chenyu You, and Bo~Wang.
\newblock Segment anything in medical images.
\newblock {\em Nature Communications}, 2024.

\bibitem{mazurowski2023segment}
Maciej~A Mazurowski, Haoyu Dong, Hanxue Gu, Jichen Yang, Nicholas Konz, and Yixin Zhang.
\newblock Segment anything model for medical image analysis: an experimental study.
\newblock {\em MedIA}, 2023.

\bibitem{LaserNet}
Gregory~P Meyer, Ankit Laddha, Eric Kee, Carlos Vallespi-Gonzalez, and Carl~K Wellington.
\newblock Lasernet: An efficient probabilistic 3d object detector for autonomous driving.
\newblock In {\em Proc. CVPR}, pages 12677--12686, 2019.

\bibitem{Nerf}
B~Mildenhall, PP~Srinivasan, M~Tancik, JT~Barron, R~Ramamoorthi, and R~Ng.
\newblock Nerf: Representing scenes as neural radiance fields for view synthesis.
\newblock In {\em Proc. ECCV}, 2020.

\bibitem{llff}
Ben Mildenhall, Pratul~P Srinivasan, Rodrigo Ortiz-Cayon, Nima~Khademi Kalantari, Ravi Ramamoorthi, Ren Ng, and Abhishek Kar.
\newblock Local light field fusion: Practical view synthesis with prescriptive sampling guidelines.
\newblock {\em ACM Transactions on Graphics (TOG)}, 38(4):1--14, 2019.

\bibitem{spin-nerf}
Ashkan Mirzaei, Tristan Aumentado-Armstrong, Konstantinos~G Derpanis, Jonathan Kelly, Marcus~A Brubaker, Igor Gilitschenski, and Alex Levinshtein.
\newblock Spin-nerf: Multiview segmentation and perceptual inpainting with neural radiance fields.
\newblock In {\em Proceedings of the IEEE/CVF Conference on Computer Vision and Pattern Recognition}, pages 20669--20679, 2023.

\bibitem{InstantNGP}
Thomas M{\"u}ller, Alex Evans, Christoph Schied, and Alexander Keller.
\newblock Instant neural graphics primitives with a multiresolution hash encoding.
\newblock {\em ACM transactions on graphics (TOG)}, 41(4):1--15, 2022.

\bibitem{donerf}
Thomas Neff, Pascal Stadlbauer, Mathias Parger, Andreas Kurz, Joerg~H Mueller, Chakravarty R~Alla Chaitanya, Anton Kaplanyan, and Markus Steinberger.
\newblock Donerf: Towards real-time rendering of compact neural radiance fields using depth oracle networks.
\newblock In {\em Computer Graphics Forum}, volume~40, pages 45--59. Wiley Online Library, 2021.

\bibitem{BEV-Seg}
Mong~H Ng, Kaahan Radia, Jianfei Chen, Dequan Wang, Ionel Gog, and Joseph~E Gonzalez.
\newblock Bev-seg: Bird's eye view semantic segmentation using geometry and semantic point cloud.
\newblock {\em arXiv preprint arXiv:2006.11436}, 2020.

\bibitem{BAEFormer}
Cong Pan, Yonghao He, Junran Peng, Qian Zhang, Wei Sui, and Zhaoxiang Zhang.
\newblock Baeformer: Bi-directional and early interaction transformers for bird's eye view semantic segmentation.
\newblock In {\em Proc. CVPR}, pages 9590--9599, 2023.

\bibitem{PointFormer}
Xuran Pan, Zhuofan Xia, Shiji Song, Li~Erran Li, and Gao Huang.
\newblock 3d object detection with pointformer.
\newblock In {\em Proc. CVPR}, pages 7463--7472, 2021.

\bibitem{peng2020large}
Junran Peng, Xingyuan Bu, Ming Sun, Zhaoxiang Zhang, Tieniu Tan, and Junjie Yan.
\newblock Large-scale object detection in the wild from imbalanced multi-labels.
\newblock In {\em Proceedings of the IEEE/CVF conference on computer vision and pattern recognition}, pages 9709--9718, 2020.

\bibitem{peng2023gaia}
Junran Peng, Qing Chang, Haoran Yin, Xingyuan Bu, Jiajun Sun, Lingxi Xie, Xiaopeng Zhang, Qi~Tian, and Zhaoxiang Zhang.
\newblock Gaia-universe: Everything is super-netify.
\newblock {\em IEEE Transactions on Pattern Analysis and Machine Intelligence}, 45(10):11856--11868, 2023.

\bibitem{peng2019efficient}
Junran Peng, Ming Sun, ZHAO-XIANG ZHANG, Tieniu Tan, and Junjie Yan.
\newblock Efficient neural architecture transformation search in channel-level for object detection.
\newblock {\em Advances in neural information processing systems}, 32, 2019.

\bibitem{peng2019pod}
Junran Peng, Ming Sun, Zhaoxiang Zhang, Tieniu Tan, and Junjie Yan.
\newblock Pod: Practical object detection with scale-sensitive network.
\newblock In {\em Proceedings of the IEEE/CVF International Conference on Computer Vision}, pages 9607--9616, 2019.

\bibitem{peng2018accelerating}
Junran Peng, Lingxi Xie, Zhaoxiang Zhang, Tieniu Tan, and Jingdong Wang.
\newblock Accelerating deep neural networks with spatial bottleneck modules.
\newblock {\em arXiv preprint arXiv:1809.02601}, 2018.

\bibitem{BEVSegFormer}
Lang Peng, Zhirong Chen, Zhangjie Fu, Pengpeng Liang, and Erkang Cheng.
\newblock Bevsegformer: Bird's eye view semantic segmentation from arbitrary camera rigs.
\newblock In {\em Proc. WACV}, pages 5935--5943, 2023.

\bibitem{tunesam1}
Zelin Peng, Zhengqin Xu, Zhilin Zeng, Lingxi Xie, Qi~Tian, and Wei Shen.
\newblock Parameter efficient fine-tuning via cross block orchestration for segment anything model.
\newblock In {\em CVPR}, 2024.

\bibitem{tunesam2}
Zelin Peng, Zhengqin Xu, Zhilin Zeng, Xiaokang Yang, and Wei Shen.
\newblock Sam-parser: Fine-tuning sam efficiently by parameter space reconstruction.
\newblock In {\em AAAI}, 2024.

\bibitem{kilonerf}
Christian Reiser, Songyou Peng, Yiyi Liao, and Andreas Geiger.
\newblock Kilonerf: Speeding up neural radiance fields with thousands of tiny mlps.
\newblock In {\em Proceedings of the IEEE/CVF international conference on computer vision}, pages 14335--14345, 2021.

\bibitem{NVOS}
Zhongzheng Ren, Aseem Agarwala, Bryan Russell, Alexander~G Schwing, and Oliver Wang.
\newblock Neural volumetric object selection.
\newblock In {\em Proceedings of the IEEE/CVF Conference on Computer Vision and Pattern Recognition}, pages 6133--6142, 2022.

\bibitem{relighting}
Viktor Rudnev, Mohamed Elgharib, William Smith, Lingjie Liu, Vladislav Golyanik, and Christian Theobalt.
\newblock Nerf for outdoor scene relighting.
\newblock In {\em European Conference on Computer Vision}, pages 615--631. Springer, 2022.

\bibitem{shao2024control4d}
Ruizhi Shao, Jingxiang Sun, Cheng Peng, Zerong Zheng, Boyao Zhou, Hongwen Zhang, and Yebin Liu.
\newblock Control4d: Efficient 4d portrait editing with text.
\newblock In {\em Proceedings of the IEEE/CVF Conference on Computer Vision and Pattern Recognition}, pages 4556--4567, 2024.

\bibitem{PointRCNN}
Shaoshuai Shi, Xiaogang Wang, and Hongsheng Li.
\newblock Pointrcnn: 3d object proposal generation and detection from point cloud.
\newblock In {\em Proc. CVPR}, pages 770--779, 2019.

\bibitem{a-nerf}
Shih-Yang Su, Frank Yu, Michael Zollh{\"o}fer, and Helge Rhodin.
\newblock A-nerf: Articulated neural radiance fields for learning human shape, appearance, and pose.
\newblock {\em Advances in neural information processing systems}, 34:12278--12291, 2021.

\bibitem{DVGO}
Cheng Sun, Min Sun, and Hwann-Tzong Chen.
\newblock Direct voxel grid optimization: Super-fast convergence for radiance fields reconstruction.
\newblock In {\em Proc. CVPR}, pages 5459--5469, 2022.

\bibitem{dvgov2}
Cheng Sun, Min Sun, and Hwann-Tzong Chen.
\newblock Improved direct voxel grid optimization for radiance fields reconstruction.
\newblock {\em arXiv preprint arXiv:2206.05085}, 2022.

\bibitem{sun20243dgstream}
Jiakai Sun, Han Jiao, Guangyuan Li, Zhanjie Zhang, Lei Zhao, and Wei Xing.
\newblock 3dgstream: On-the-fly training of 3d gaussians for efficient streaming of photo-realistic free-viewpoint videos.
\newblock In {\em Proceedings of the IEEE/CVF Conference on Computer Vision and Pattern Recognition}, pages 20675--20685, 2024.

\bibitem{tang2023dreamgaussian}
Jiaxiang Tang, Jiawei Ren, Hang Zhou, Ziwei Liu, and Gang Zeng.
\newblock Dreamgaussian: Generative gaussian splatting for efficient 3d content creation.
\newblock {\em arXiv preprint arXiv:2309.16653}, 2023.

\bibitem{tang2023can}
Lv~Tang, Haoke Xiao, and Bo~Li.
\newblock Can sam segment anything? when sam meets camouflaged object detection.
\newblock {\em arXiv preprint arXiv:2304.04709}, 2023.

\bibitem{N3F}
Vadim Tschernezki, Iro Laina, Diane Larlus, and Andrea Vedaldi.
\newblock Neural feature fusion fields: 3d distillation of self-supervised 2d image representations.
\newblock In {\em 2022 International Conference on 3D Vision (3DV)}, pages 443--453. IEEE, 2022.

\bibitem{wang2021uncertainty}
Yuxi Wang, Junran Peng, and ZhaoXiang Zhang.
\newblock Uncertainty-aware pseudo label refinery for domain adaptive semantic segmentation.
\newblock In {\em Proceedings of the IEEE/CVF international conference on computer vision}, pages 9092--9101, 2021.

\bibitem{humannerf}
Chung-Yi Weng, Brian Curless, Pratul~P Srinivasan, Jonathan~T Barron, and Ira Kemelmacher-Shlizerman.
\newblock Humannerf: Free-viewpoint rendering of moving people from monocular video.
\newblock In {\em Proceedings of the IEEE/CVF conference on computer vision and pattern Recognition}, pages 16210--16220, 2022.

\bibitem{wewer2024latentsplat}
Christopher Wewer, Kevin Raj, Eddy Ilg, Bernt Schiele, and Jan~Eric Lenssen.
\newblock latentsplat: Autoencoding variational gaussians for fast generalizable 3d reconstruction.
\newblock {\em arXiv preprint arXiv:2403.16292}, 2024.

\bibitem{wu20244d}
Guanjun Wu, Taoran Yi, Jiemin Fang, Lingxi Xie, Xiaopeng Zhang, Wei Wei, Wenyu Liu, Qi~Tian, and Xinggang Wang.
\newblock 4d gaussian splatting for real-time dynamic scene rendering.
\newblock In {\em Proceedings of the IEEE/CVF Conference on Computer Vision and Pattern Recognition}, pages 20310--20320, 2024.

\bibitem{wu2023medical}
Junde Wu, Rao Fu, Huihui Fang, Yuanpei Liu, Zhaowei Wang, Yanwu Xu, Yueming Jin, and Tal Arbel.
\newblock Medical sam adapter: Adapting segment anything model for medical image segmentation.
\newblock {\em arXiv preprint arXiv:2304.12620}, 2023.

\bibitem{xu2024agg}
Dejia Xu, Ye~Yuan, Morteza Mardani, Sifei Liu, Jiaming Song, Zhangyang Wang, and Arash Vahdat.
\newblock Agg: Amortized generative 3d gaussians for single image to 3d.
\newblock {\em arXiv preprint arXiv:2401.04099}, 2024.

\bibitem{yang2024gaussianobject}
Chen Yang, Sikuang Li, Jiemin Fang, Ruofan Liang, Lingxi Xie, Xiaopeng Zhang, Wei Shen, and Qi~Tian.
\newblock Gaussianobject: Just taking four images to get a high-quality 3d object with gaussian splatting.
\newblock {\em arXiv preprint arXiv:2402.10259}, 2024.

\bibitem{BEVFormerv2}
Chenyu Yang, Yuntao Chen, Hao Tian, Chenxin Tao, Xizhou Zhu, Zhaoxiang Zhang, Gao Huang, Hongyang Li, Yu~Qiao, Lewei Lu, et~al.
\newblock Bevformer v2: Adapting modern image backbones to bird's-eye-view recognition via perspective supervision.
\newblock In {\em Proc. CVPR}, pages 17830--17839, 2023.

\bibitem{gaussian-grouping}
Mingqiao Ye, Martin Danelljan, Fisher Yu, and Lei Ke.
\newblock Gaussian grouping: Segment and edit anything in 3d scenes.
\newblock {\em arXiv preprint arXiv:2312.00732}, 2023.

\bibitem{yi2023gaussiandreamer}
Taoran Yi, Jiemin Fang, Guanjun Wu, Lingxi Xie, Xiaopeng Zhang, Wenyu Liu, Qi~Tian, and Xinggang Wang.
\newblock Gaussiandreamer: Fast generation from text to 3d gaussian splatting with point cloud priors.
\newblock {\em arXiv preprint arXiv:2310.08529}, 2023.

\bibitem{plenoctrees}
Alex Yu, Ruilong Li, Matthew Tancik, Hao Li, Ren Ng, and Angjoo Kanazawa.
\newblock Plenoctrees for real-time rendering of neural radiance fields.
\newblock In {\em Proceedings of the IEEE/CVF International Conference on Computer Vision}, pages 5752--5761, 2021.

\bibitem{yu2024cogs}
Heng Yu, Joel Julin, Zolt{\'a}n~{\'A} Milacski, Koichiro Niinuma, and L{\'a}szl{\'o}~A Jeni.
\newblock Cogs: Controllable gaussian splatting.
\newblock In {\em Proceedings of the IEEE/CVF Conference on Computer Vision and Pattern Recognition}, pages 21624--21633, 2024.

\bibitem{yu2023inpaint}
Tao Yu, Runseng Feng, Ruoyu Feng, Jinming Liu, Xin Jin, Wenjun Zeng, and Zhibo Chen.
\newblock Inpaint anything: Segment anything meets image inpainting.
\newblock {\em arXiv preprint arXiv:2304.06790}, 2023.

\bibitem{gof}
Zehao Yu, Torsten Sattler, and Andreas Geiger.
\newblock Gaussian opacity fields: Efficient and compact surface reconstruction in unbounded scenes.
\newblock {\em arXiv preprint arXiv:2404.10772}, 2024.

\bibitem{zeng2024stag4d}
Yifei Zeng, Yanqin Jiang, Siyu Zhu, Yuanxun Lu, Youtian Lin, Hao Zhu, Weiming Hu, Xun Cao, and Yao Yao.
\newblock Stag4d: Spatial-temporal anchored generative 4d gaussians.
\newblock {\em arXiv preprint arXiv:2403.14939}, 2024.

\bibitem{mobileSAM}
Chaoning Zhang, Dongshen Han, Yu~Qiao, Jung~Uk Kim, Sung-Ho Bae, Seungkyu Lee, and Choong~Seon Hong.
\newblock Faster segment anything: Towards lightweight sam for mobile applications.
\newblock {\em arXiv preprint arXiv:2306.14289}, 2023.

\bibitem{zhang2023sam3d}
Dingyuan Zhang, Dingkang Liang, Hongcheng Yang, Zhikang Zou, Xiaoqing Ye, Zhe Liu, and Xiang Bai.
\newblock Sam3d: Zero-shot 3d object detection via segment anything model.
\newblock {\em SCIS}, 2023.

\bibitem{zhang2025general}
Guowen Zhang, Junsong Fan, Liyi Chen, Zhaoxiang Zhang, Zhen Lei, and Lei Zhang.
\newblock General geometry-aware weakly supervised 3d object detection.
\newblock In {\em European Conference on Computer Vision}, pages 290--309. Springer, 2025.

\bibitem{zhang2024voxel}
Guowen Zhang, Lue Fan, Chenhang He, Zhen Lei, Zhaoxiang Zhang, and Lei Zhang.
\newblock Voxel mamba: Group-free state space models for point cloud based 3d object detection.
\newblock {\em arXiv preprint arXiv:2406.10700}, 2024.

\bibitem{zhang2024cityx}
Shougao Zhang, Mengqi Zhou, Yuxi Wang, Chuanchen Luo, Rongyu Wang, Yiwei Li, Xucheng Yin, Zhaoxiang Zhang, and Junran Peng.
\newblock Cityx: Controllable procedural content generation for unbounded 3d cities.
\newblock {\em arXiv preprint arXiv:2407.17572}, 2024.

\bibitem{zhang2021knet}
Wenwei Zhang, Jiangmiao Pang, Kai Chen, and Chen~Change Loy.
\newblock K-net: Towards unified image segmentation.
\newblock {\em Advances in Neural Information Processing Systems}, 34:10326--10338, 2021.

\bibitem{nerfactor}
Xiuming Zhang, Pratul~P Srinivasan, Boyang Deng, Paul Debevec, William~T Freeman, and Jonathan~T Barron.
\newblock Nerfactor: Neural factorization of shape and reflectance under an unknown illumination.
\newblock {\em ACM Transactions on Graphics (ToG)}, 40(6):1--18, 2021.

\bibitem{zhang2022delving}
Zhaoxiang Zhang, Cong Pan, and Junran Peng.
\newblock Delving into the effectiveness of receptive fields: Learning scale-transferrable architectures for practical object detection.
\newblock {\em International Journal of Computer Vision}, 130(4):970--989, 2022.

\bibitem{fastsam}
Xu~Zhao, Wenchao Ding, Yongqi An, Yinglong Du, Tao Yu, Min Li, Ming Tang, and Jinqiao Wang.
\newblock Fast segment anything.
\newblock {\em arXiv preprint arXiv:2306.12156}, 2023.

\bibitem{semantic-nerf}
Shuaifeng Zhi, Tristan Laidlow, Stefan Leutenegger, and Andrew~J Davison.
\newblock In-place scene labelling and understanding with implicit scene representation.
\newblock In {\em Proceedings of the IEEE/CVF International Conference on Computer Vision}, pages 15838--15847, 2021.

\bibitem{zhou2024scenex}
Mengqi Zhou, Yuxi Wang, Jun Hou, Chuanchen Luo, Zhaoxiang Zhang, and Junran Peng.
\newblock Scenex: Procedural controllable large-scale scene generation via large-language models.
\newblock {\em arXiv preprint arXiv:2403.15698}, 2024.

\bibitem{featureGS}
Shijie Zhou, Haoran Chang, Sicheng Jiang, Zhiwen Fan, Zehao Zhu, Dejia Xu, Pradyumna Chari, Suya You, Zhangyang Wang, and Achuta Kadambi.
\newblock Feature 3dgs: Supercharging 3d gaussian splatting to enable distilled feature fields.
\newblock {\em arXiv preprint arXiv:2312.03203}, 2023.

\bibitem{zhou2024dreamscene360}
Shijie Zhou, Zhiwen Fan, Dejia Xu, Haoran Chang, Pradyumna Chari, Tejas Bharadwaj, Suya You, Zhangyang Wang, and Achuta Kadambi.
\newblock Dreamscene360: Unconstrained text-to-3d scene generation with panoramic gaussian splatting.
\newblock {\em arXiv preprint arXiv:2404.06903}, 2024.

\bibitem{zhou2024drivinggaussian}
Xiaoyu Zhou, Zhiwei Lin, Xiaojun Shan, Yongtao Wang, Deqing Sun, and Ming-Hsuan Yang.
\newblock Drivinggaussian: Composite gaussian splatting for surrounding dynamic autonomous driving scenes.
\newblock In {\em Proceedings of the IEEE/CVF Conference on Computer Vision and Pattern Recognition}, pages 21634--21643, 2024.

\bibitem{ewa}
Matthias Zwicker, Hanspeter Pfister, Jeroen Van~Baar, and Markus Gross.
\newblock Ewa volume splatting.
\newblock In {\em Proceedings Visualization, 2001. VIS'01.}, pages 29--538. IEEE, 2001.

\end{thebibliography}
}

\newpage

\vfill

\end{document}